\pdfoutput=1

\documentclass[11pt]{article}

\usepackage[final]{acl}

\usepackage{times}
\usepackage{latexsym}

\usepackage[T1]{fontenc}

\usepackage[utf8]{inputenc}

\usepackage{microtype}

\usepackage{inconsolata}
\usepackage{soul}
\usepackage{graphicx}
\usepackage[export]{adjustbox}
\usepackage{subcaption}
\usepackage{multirow}
\usepackage{xcolor}
\usepackage{longtable}
\usepackage{flafter}
\usepackage{authblk}
\usepackage{float}
\newcommand{\hltext}[3][RGB]{%
  \begingroup
  \definecolor{hlcolor}{#1}{#2}\sethlcolor{hlcolor}%
  \hl{#3}%
  \endgroup
}

\title{Communication is All You Need:\\Persuasion Dataset Construction via Multi-LLM Communication}

\author[1]{\textbf{Weicheng Ma}}
\author[2]{\textbf{Hefan Zhang}}
\author[2]{\textbf{Ivory Yang}}
\author[3]{\textbf{Shiyu Ji}}
\author[3]{\authorcr \textbf{Joice Chen}}
\author[3]{\textbf{Farnoosh Hashemi}}
\author[3]{\textbf{Shubham Mohole}}
\author[2]{\textbf{Ethan Gearey}}
\author[3]{\authorcr \textbf{Michael Macy}}
\author[2]{\textbf{Saeed Hassanpour}}
\author[2]{\textbf{Soroush Vosoughi}}
\affil[1]{College of Computing, Georgia Institute of Technology}
\affil[2]{Department of Computer Science, Dartmouth College}
\affil[3]{Cornell University}
\affil[1]{\texttt{wma76@gatech.edu}}

\begin{document}
\maketitle
\begin{abstract}
 
Large Language Models (LLMs) have shown proficiency in generating persuasive dialogue, yet concerns about the fluency and sophistication of their outputs persist. This paper presents a multi-LLM communication framework designed to enhance the generation of persuasive data automatically. This framework facilitates the efficient production of high-quality, diverse linguistic content with minimal human oversight. Through extensive evaluations, we demonstrate that the generated data excels in naturalness, linguistic diversity, and the strategic use of persuasion, even in complex scenarios involving social taboos. The framework also proves adept at generalizing across novel contexts. Our results highlight the framework's potential to significantly advance research in both computational and social science domains concerning persuasive communication.

\end{abstract}
 
\section{Introduction}
Persuasion techniques play a critical role in shaping societal behaviors and public opinion \cite{fogg2009behavior, braca2023developing}, which has led to sustained interest across a range of disciplines. Social science research has established detailed taxonomies of persuasion strategies \cite{shrum2012persuasion, lukin2017argument}, while datasets have been developed to cover various domains, including charitable donations \cite{wang2019persuasion}, argument ranking in debates \cite{toledo2019automatic}, detecting mental manipulation \cite{wang-etal-2024-mentalmanip, yang2024enhanced}, and understanding advertising strategies \cite{kumar2023persuasion}. Despite these advances, ambiguities persist in defining persuasion \cite{pauli2022modelling}, and applying persuasion strategies across different contexts remains complex \cite{bai2021m2p2, schaefer2023towards, piskorski2023multilingual}. Additionally, the high cost of manually annotating quality data poses a significant challenge \cite{lai2022werewolf}.

The advent of large language models (LLMs) has unlocked new possibilities for enhancing various forms of communication, including online political discourse \cite{argyle2023leveraging, bai2023artificial}, personalized advertising \cite{matz2024potential, meguellati2024good}, public health messaging \cite{lim2023artificial, espinosa2024use}, and opinion shaping on social media \cite{meier2024llm}. Recent research, such as that by \citet{persugpt}, has begun exploring LLM-generated persuasive dialogues. However, their approach is limited to simple, two-party dialogues where a persuader seeks to change the persuadee’s viewpoint. These dialogues often lack depth, presenting brief exchanges with simplistic logic and unnatural flow, restricting their usefulness for studying persuasion in more complex settings.

In response to these limitations, we propose a multi-agent framework for generating persuasion data. In this framework, multiple agents are assigned distinct roles, ensuring that each aspect of the dialogue generation process is handled efficiently. This structure minimizes the risk of an agent missing important details due to task abstraction or prompt complexity, a common issue in LLM prompting \cite{prompt-complexity-1, prompt-complexity-2}. Additionally, auxiliary agents manage dialogue flow to ensure that the resulting exchanges are coherent, logically consistent, and incorporate diverse persuasive strategies, simulating natural human conversation.
Our approach imposes no preconditions regarding speakers, language styles, domains, or persuasion strategies, allowing it to generate a wide range of dialogues. For instance, our framework can support adversarial dialogues, where both participants attempt to persuade one another while maintaining their original positions. Moreover, we employ a continuous labeling scheme to measure the degree of perspective change throughout the dialogue, avoiding the limitations of binary utterance labels. This framework also integrates ethical considerations, incorporating cultural norms and taboos from NormBank \cite{ziems2023normbank} to explore ethically challenging persuasive scenarios, such as dialogues involving manipulation or unethical persuasion.

Careful analyses conducted by experts from both NLP and social sciences confirm the quality of our generated dialogues, particularly in terms of their naturalness, logical structure, and diversity of persuasion strategies. Our sentence-level persuasiveness labels align closely with human judgments (see Appendix \ref{appendix:label-validation}).

We further demonstrate the flexibility of our framework through experiments controlling for specific persuasion strategies and in more complex scenarios, such as multi-party conversations. Across all tested configurations, our framework consistently produced high-quality dialogues, showcasing its adaptability and generalizability.
These findings indicate that our framework offers a robust platform for studying persuasion techniques, particularly in high-stakes contexts where ethical concerns, such as the spread of misinformation and propaganda, are paramount \cite{chen2023combating, jones2024detecting}.

\begin{figure}[t]
    \centering
    \includegraphics[width=1\linewidth]{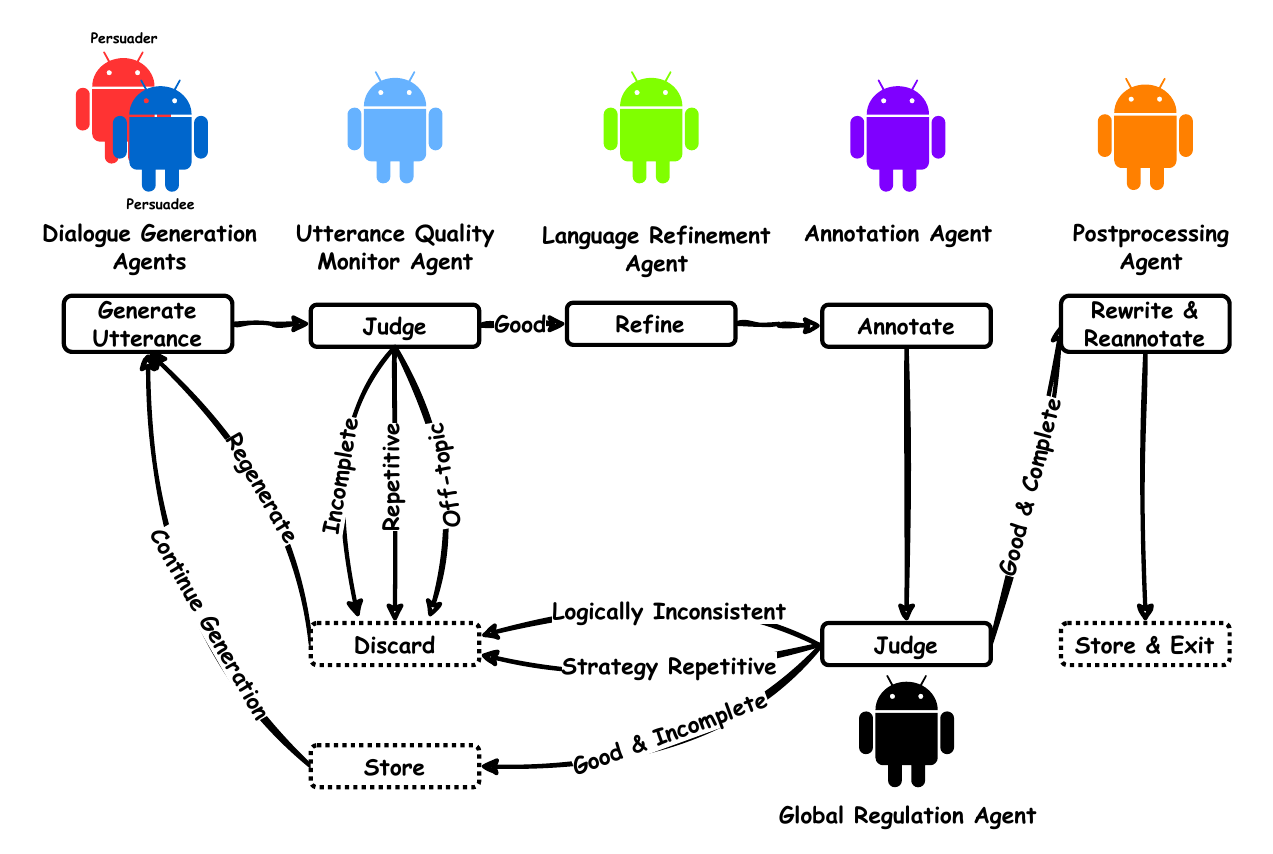}
\caption{Overview of our data generation and annotation framework. Prior to dialogue generation, each agent is assigned specific tasks and given predefined stances to maintain throughout the conversation.}
    \label{fig:annotation-framework}
    \vspace{-5mm}
\end{figure}

\section{Multi-Agent Data Generation \& Annotation Framework}
Our framework incorporates 6 groups of language agents as shown in Figure \ref{fig:annotation-framework}. In our experiments, all agents utilize a GPT-3.5 backbone, except for the utterance quality monitor and global regulation agents which are based on GPT-4 due to their need for advanced reasoning capabilities and enhanced memory retention. Note that this choice of LLMs aims to balance data generation costs with quality, and using more powerful models could further improve the effectiveness of our approach.
Our preliminary experiments on model selection are outlined in Appendix \ref{appendix:model-selection}.

\subsection{Dialogue Generation Agents}
We adopt a methodology for generating multi-round conversations by cyclically using the output from one language agent as the input for another \cite{small-town}. This technique has been validated to produce extended, logically consistent dialogues that fulfill our project requirements.

Our framework initializes the generative agents with a description of the task settings, the predefined tasks for each language agent, and guidelines governing the models' generations, as illustrated in Figure \ref{fig:system-persuader-persuadee}.
The task choices for each agent are not constrained, for instance, drawing on a cultural taboo that ``one should not pick flowers in a cemetery'' from NormBank, we could challenge the persuader to convince the persuadee to pick flowers in a cemetery, while the persuadee is instructed to resist and, if possible, persuade the persuader to abandon such thoughts.

The dialogues commence when we prompt a persuader agent with ``Start the conversation.'' This setup initiates a structured yet dynamic interaction between the speakers, allowing us to closely observe and analyze their persuasive strategies.

\subsection{Utterance Quality Monitor Agent}
Due to the inherent limitations of LLMs, dialogue generation agents may occasionally produce incomplete, repetitive, or off-topic content. To address these issues, we introduce a specialized LLM agent responsible for tracking the persuasion topic and generation history to evaluate new generations.

The initialization prompt of the utterance quality monitor agent is shown in Figure \ref{fig:system-language-utterance-monitor}.
During dialogue generation, this agent inspects every new utterance to check if they \ul{ends unexpectedly}, \ul{repeats a previous utterance}, or \ul{goes off the topic of the dialogue} in a sequence.
If an utterance is red-flagged for any issue, the author agent is requested to revise the utterance based on the diagnoses. Otherwise, before proceeding to the next utterance, the utterance quality monitor agent is prompted to update its memory, storing the reviewed utterance for future judgments.

\subsection{Language Refinement Agent}
Raw text produced by dialogue generation agents often adopts a conclusive rather than conversational tone, primarily because the agents are prompted in a question-answering format. This could lead to stylistic conflicts with surrounding utterances. Additionally, the generations frequently include tone-softening phrases like ``I understand your concerns,'' or unnecessary affirmations such as agreeing with the other speaker's views, which dilute the strength of arguments. Over the course of the conversation, these issues can compound, leading to dialogues dominated by language softeners and lacking in persuasive content.

To address this issue,
we adopt a language refinement agent tasked with stripping out polite but superfluous phrases, thereby sharpening the dialogue's focus on substantive content. System message to this agent is shown in Figure \ref{fig:system-language-refine}. 2 examples are also provided to the agent to further regulate its behaviors.
Subsequent operations, including continued dialogue generation and persuasiveness labeling, are predicated on the output from the language refinement agent, ensuring that the conversation maintains its relevance and effectiveness in conveying persuasive arguments.

\subsection{Persuasiveness Annotation Agent}
After generating each round of conversation, we employ a persuasiveness annotation agent to assess the extent of perspective shifts in each speaker, assigning a score ranging from 0 to 1. 
Figure \ref{fig:system-annotation} illustrates the system message fed to the persuasiveness annotation agent before the generation starts.
In practice, we provide the annotation agent with two scoring examples to guide its behavior and minimize scoring errors, such as incorrectly assigning a score of 1 to a conversation round with no perspective shifts (Figure \ref{fig:25}). Note that these scores reflect the cumulative viewpoint shifts across all prior rounds of communication, facilitating the analysis of gradual persuasion rather than focusing solely on the impact of a single utterance.

\subsection{Global Regulation Agent}
We employ a global regulation agent to ensure smooth logical flow in the generated conversations and to determine the appropriate time to conclude the dialogue. The system message to the global regulation agent is depicted in Figure \ref{fig:system-regulation}.

After each round of utterances is generated and annotated, we prompt this agent to verify whether any changes in each speaker's perspectives are logically influenced by the preceding utterance and whether the newly generated utterances avoid repeating previously used strategies within the same conversation. If the logical connections are insufficient or no new persuasive attempts are made, the dialogue generation agents are asked to revise their responses based on feedback from the global regulation agent. 
Once the revised generation passes these checks, the agent’s internal memory is updated accordingly.
Then the agent is prompted to assess whether the speakers have reached a mutual agreement or if no new information is likely to be introduced next, indicating that the dialogue should be concluded. Although the ideal conclusion involves the persuader and persuadee agreeing on the preset task, conversations can often devolve into repetitive and unproductive arguments
(Figure \ref{fig:112}) \cite{xu2022learning}. To prevent such stagnation, we allow the dialogue to conclude even if complete agreement is not reached. The global regulation agent is responsible for determining when to end the dialogue, at which point the conversation is terminated and the agent's memory is reset.

\subsection{Postprocessing Agent}
After generating and annotating a full dialogue, we use a postprocessing agent to enhance content smoothness and naturalness. As shown in Figure \ref{fig:system-postprocessing}, the agent removes redundant language, improves logical flow, and enhances language diversity. It also merges labels and reassigns them to modified utterances if the number of dialogue rounds changes.


\begin{table*}[!h]
\centering
\small
\resizebox{\textwidth}{!}{
\begin{tabular}{p{5cm}p{8cm}p{8cm}}
\hline
\textbf{Annotator Agreement Type} & \textbf{Model-generated Utterances} & \textbf{Utterances with Human Rewrites}\\
\hline
\multirow{3}{=}{\textbf{Human annotators disagree with each other (49\% of the sample)}} & Despite your persuasive argument, I maintain my decision to forgo shoes in the library. I am convinced that personal comfort and cleanliness can coexist without conforming to the practice of wearing shoes indoors. & Even with your compelling argument, I stand by my choice to go without shoes in the library. I believe that personal comfort and cleanliness can coexist without adhering to the norm of wearing shoes indoors. \\\cline{2-3}
& Undeniably, caution is essential, yet many individuals choose their offerings with respect and adherence to guidelines. With proper education and awareness, we can continue this tradition in a way that respects both the site and the environment. Striking a balance between honoring history and preserving nature is key. & Many individuals choose their offerings with respect and adherence to guidelines while keeping caution in the back of their mind though it is essential. Striking a balance between honoring history and preserving nature is key. With proper education and awareness, we can continue this tradition in a way that respects both the site and the environment. \\\cline{2-3}
& I respect your decision, but I urge you to consider the potential hazards. Your well-being is vital, and alternative relaxation methods could offer similar benefits without risk. Balancing peace of mind with safety is crucial for healthy self-care practices. & Though I respect your decision, I advise you to familiarize yourself with the potential hazards of this practice. Your well-being is vital, and alternative relaxation methods could offer similar benefits without risk. Balancing peace of mind with safety is crucial for healthy self-care practices.\\\cline{2-3}
\hline
\multirow{3}{=}{\textbf{Both human annotators correctly identified LLM-generated messages (29.25\% of the sample)}} & Despite your precautions, the risks of encountering dangerous animals or losing your way in the dark are substantial. Our senses are less effective at night, making it difficult to respond quickly to threats. It's safer to err on the side of caution and avoid the forest after dark. & The risks of encountering dangerous animals or losing your way in the dark are substantial regardless of your preparations and whatever precautions you take. Our senses are less effective at night, making it difficult to respond quickly to threats. It's safer to be cautious and avoid the forest after dark. \\\cline{2-3}
& If there's any other way I can help you, please reach out. Enjoy your day at the bakery. & Thank you for stopping by our bakery; if there are any questions, I am glad to help.\\\cline{2-3}
& I trust in the power of spiritual intervention to bring about the positive changes needed in my apartment. Would you reconsider and assist me? &  I believe that powerful spiritual intervention can bring the positive changes my apartment needs. Would you change your mind and help me? \\\cline{2-3}
\hline
\multirow{3}{=}{\textbf{Both human annotators took human-written messages as model-generated messages (21.75\% of the sample)}} & While I understand the value of learning to provide basic medications under guidance, I still believe that it's crucial to ensure the animals' safety and well-being by relying on trained professionals for medication administration. I will continue to focus on supporting the animals' welfare through means that align with my capabilities, while prioritizing their health and security. &  The value of learning to provide basic medications under guidance is undeniable, but I still believe that it's crucial to ensure the animals' safety and well-being by relying on trained professionals for medication administration. I will continue to focus on supporting the animals' welfare through means that align with my capabilities while prioritizing their health and security.\\\cline{2-3}
&Wearing high heels doesn't necessarily mean compromising comfort and safety. You can opt for comfortable, stable high heel boots specifically designed for outdoor use. They offer style and functionality, allowing you to experience the thrill of skiing or snowboarding while looking chic. It's a chance to redefine traditional ski resort attire and make striking fashion statements. & High heels does not mean that safety and comfort are compromized, rather there are options like staboe high heel boots aimed for outdoor use and they offer style and functionality. You can experience the thrill of skiing or snowboarding while looking chic and its a opportunity to redefine traditional ski resort attire making striking fashion choices. \\\cline{2-3}
& While I respect your perspective, I feel comfortable with my decision. I enjoy the freedom and comfort of skiing without a helmet. My cautious approach and awareness help me ensure my safety. & I value your opinion, but I feel confident in my decision. I take pleasure in the freedom and comfort of skiing without a helmet. My mindful approach and attentiveness enable me to keep myself safe. \\
\hline

\end{tabular}
}

\caption{Examples of Annotator Agreement on Utterance-Level Human v.s. LLM Differentiation Task. Across 400 sampled pairs of utterances, annotators disagreed on which sentence was LLM-generated in 49\% of cases. In 29.25\% of the pairs, both annotators successfully identified the LLM-generated language, while in 21.75\% of the pairs, neither annotator was able to detect the LLM-generated language.}
\label{table:utterance_example}
\end{table*}

\section{Data Quality Assessment} \label{sct:analyses}
To evaluate our data generation framework, we constructed a small dataset of 200 dialogues using randomly selected norms from NormBank for human validation. These norms consist of 98 taboos, 76 normal behaviors, and 26 expected behaviors. We intentionally placed greater emphasis on taboos because these behaviors often conflict with widely accepted moral standards, causing LLMs to refuse to generate persuasive dialogues (Figure \ref{fig:exp-llm-refusal-taboo}). 
As such, they present a unique challenge in persuasion scenarios for both humans and LLMs.

Our data assessment plan focuses on three key aspects, progressing from more specific to broader levels of analysis: (1) the \ul{language fluency} of individual utterances, (2) the \ul{the topic, semantic, and logical coherence} of entire conversations, and (3) the \ul{language and strategy diversity} of conversations generated under the same topic and context.

\subsection{Utterance-Level Quality Assessment} \label{sct:utterance-level-assessment}
A critical goal for our framework is that each generated utterance should closely resemble a human-written sentence. To validate this, we 
conduct (a) a quantitative annotation task to differentiate between model-generated sentences and human-rewritten sentences, followed by (b) a qualitative error analysis that combines annotator feedback with insights from an LLM on sentences that multiple annotators agreed were distinguishable.

\subsubsection{Quantitative Differentiation Task}
The differentiation task aims to assess how accurately human annotators could tell model-generated sentences apart from those rewritten by humans. Similar tasks have been discussed in \citet{gehrmann-etal-2019-gltr}, \citet{ippolito-etal-2020-automatic} and \citet{Ma2023AIVH}. For our evaluation, we obtained a stratified sample of 400 utterances from 150 random sample dialogues to ensure equal representation of utterances from both the persuader and persuadee agents, covering different rounds of persuasion to reflect the dataset distribution.

\noindent\textbf{Manual Rewriting.} 2 native English speakers were asked to rewrite each sampled utterance to provide reference texts that model-generated utterances will be compared against. Each assistant was assigned 200 utterances. As shown in Figure \ref{fig:instr_rewrite}, they were instructed to retain the original meaning while improving clarity, grammar, and natural phrasing. Additionally, they could refine any awkward or unclear phrasing without altering the intended message. 



\begin{table*}
\centering
\small
\resizebox{\textwidth}{!}{
\begin{tabular}{p{2.5cm}p{8cm}p{4.8cm}p{2cm}p{1cm}p{2cm}}
\hline
\textbf{Category} & \textbf{Statements} & \textbf{Related Work} & \textbf{Avg. Score} & \textbf{$\kappa$} & \textbf{Weighted $\kappa$}\\
\hline
\multirow{6}{=}{\textbf{Interpersonal Responses}} & \textbf{\emph{Coherence}} & & & &  \\
 & The speakers respond logically to the immediate conversation.
 & \citet{ke-etal-2018-generating}, \citet{wu-etal-2019-proactive}, \citet{liang2021} & 2.969 & 0.657  & 0.657 \\
 & The arguments makes sense given its context. & \citet{zhu-etal-2019-retrieval} & 2.653 & 0.473  & 0.481 \\
 & \textbf{\emph{Informativeness}} &  & &  & \\
 & The utterances build on prior information in near context.
 & \citet{moghe-etal-2018-towards}, 
 \citet{Young_Cambria_Chaturvedi_Zhou_Biswas_Huang_2018}, \citet{lin-etal-2019-moel}, \citet{wu-etal-2019-proactive}  & 2.755 & 0.339  & 0.339 \\
 & The utterances introduce relevant new information or arguments. & \citet{ke-etal-2018-generating}, \citet{wu-etal-2019-proactive}, \citet{zhu-etal-2019-retrieval} & 2.337 & 0.410  & 0.459 \\\hline
\multirow{2}{=}{\textbf{Overall Fluency}} & The arguments overall are communicated clearly.
 & \citet{moghe-etal-2018-towards}, \citet{lin-etal-2019-moel} & 3 & NA  & NA \\
 & The conversation sounds human-like and fluent overall. & \citet{ke-etal-2018-generating}, \citet{wu-etal-2019-proactive}, \citet{zhu-etal-2019-retrieval}, \citet{ji-etal-2022-achieving} & 2.561 & 0.557  & 0.576 \\
\hline
\textbf{Internal Role Consistency}& There are no sudden shifts in a speaker’s objectives or stance without a clear explanation. & \citet{moghe-etal-2018-towards}, \citet{ji-etal-2022-achieving} & 2.765 & 0.397 & 0.546
\\\hline
\textbf{Topic Consistency} & The conversation stays on topic & \citet{moghe-etal-2018-towards}, \citet{ji-etal-2022-achieving} & 2.878 & 0.548 & 0.645
\\
\hline
\end{tabular}
}

\caption{Dialogue-level Quality Evaluation. 2 annotators assessed 50 randomly selected dialogues on the criteria listed above, using a likert scale of 1 - Not Accurate, 2 - Somewhat Accurate, and 3 - Accurate. We report the average scores across all dialogues for each measured dimension. Both linearly weighted (Weighted $\kappa$) and unweighted ($\kappa$) inter-rater consistency scores are calculated, with all results showing significant agreement. 
}
\label{table:dialogue_evaluation}
\vspace{-3mm}
\end{table*}


\noindent\textbf{Human Validation.} After manual rewriting, we created a dataset consisting of pairs of model-generated utterances and their corresponding rewritten versions. 3 fluent English-speaking annotators were then tasked with identifying the model-generated utterance in each pair. The instructions provided to the annotators are shown in Figure \ref{fig:instr_human_val}.


    
    


Each utterance in the dataset was annotated by 2 annotators, and annotators were encouraged to comment on examples they found interesting. Note that if the annotators were unable to distinguish between sentences and resorted to random guessing, the expected accuracy for both annotators correctly identifying model-generated utterances would be 25\%. Comparing the actual accuracy to this baseline helps determine whether the model-generated utterances appeared natural to the annotators.

Of the 400 utterance pairs, the model-generated utterances in 117 pairs (29.25\%) were correctly identified by both annotators, slightly going above the random baseline of 25\%. In 49\% cases (98 pairs), the annotators disagreed, and in 21.75\% cases, both annotators resulted in incorrect identifications.
Individual annotator accuracies were 0.546, 0.558, and 0.508. 
The results are close to random guessing, suggesting the challenge of distinguishing utterances generated by our framework from human-written ones.
Example utterance pairs and their annotator labels are provided in Table \ref{table:utterance_example}.

\subsubsection{Model-assisted Error Analysis}
The quantitative findings indicate that our framework generally produces high-quality utterances nearly indistinguishable from human-written sentences. To follow up, we conducted a qualitative error analysis on the samples correctly distinguished by both annotators to identify areas for improvement. Precisely, all 117 such utterances were submitted to OpenAI's o1-preview model \cite{OpenAI_2024} for further analysis to understand reasons behind their distinguishability by humans. The prompt for this task is shown in Figure \ref{fig:instr_llm_val}.









Out of the 117 pairs, o1-preview correctly distinguished 72 pairs (61.2\%), suggesting that even for LLMs, utterances generated by our framework are close to human writings.
As suggested by o1-preview and verified by human annotators (Table \ref{table:utterance_error}), major causes of unnaturalness (frequencies cited in parentheses) in these 72 less human-like utterances include overly formal language or detached tone or word use (88.9\%), lengthy sentences, redundancy, verbosity, and repetition (68.1\%), unnatural syntax, word choice, and language style (58.3\%), complex sentence structures (34.7\%), use of generic words and cliché phrases (23.6\%), overly perfect grammar (13.9\%), and LLM-style closing phrases (12.5\%). 

\subsection{Dialogue Smoothness and Naturalness} \label{sct:analyses:naturalness}
We further conduct dialogue-level analyses on our sample data to ensure that each generated dialogue is logically coherent and effective in persuasion.

\subsubsection{Dialogue Quality Annotation}

We first developed a systematic rubric for evaluating the overall quality of persuasive dialogues.
Our evaluation is conducted on (a) the local level, which examines each argument-response pair between the speakers, and (b) the global level, which considers the conversation as a whole. Evaluations are based on existing human evaluation dimensions for open dialogue systems and emphasize three key aspects: the interaction between persuader and persuadee, the consistency of individual participants across multiple rounds, and the alignment of utterances with the topic. Detailed criteria and their references are outlined in Table \ref{table:dialogue_evaluation}.


\begin{table*}[t]
\centering
\small
\resizebox{\textwidth}{!}{
\begin{tabular}{p{2.3cm}p{3.2cm}p{9.5cm}p{3.2cm}}
\hline
\textbf{Category} & \textbf{Technique} & \textbf{Definition} & \textbf{Citation}\\
\hline
\multirow{2}{=}{\textbf{External Validity}} & \textbf{Popularity} &  Invoke popular opinion or experiences as support for uptake & \multirow{5}{=}{\citealp{anand_2011}, \citealp{iyer_2019}} \\\cline{2-3}
 & \textbf{Authority} & Appeals to authority (bosses, experts, trend-setters). &  \\\cline{1-3}
\multirow{2}{=}{\textbf{Outcomes}} & \textbf{Outcomes} & Mentions some particular consequences from up take or failure to up take & \\\cline{2-3}
 & \textbf{Threat/Promise} & Poses direct threat or promise to the persuadee & \\\cline{1-3}
\textbf{Generalizations} & \textbf{Deontic/Moral Appeals} & Mentions duties or obligations, moral goodness, badness & \\\cline{1-3}
\multirow{2}{=}{\textbf{Interpersonal}} & \textbf{Empathy} & Attempts to make the persuadee connect with someone else’s emotional perspective &   \\\cline{2-4}
 & \textbf{Scarcity} & Mentions rarity, urgency, or opportunity of some outcome & \multirow{2}{=}{\citealp{anand_2011},  \citealp{iyer_2019}, \citealp{kumar2023persuasion}} \\\cline{1-3}
 \multirow{2}{=}{\textbf{Other}} & \textbf{Logical Appeal} & Provides a justification for an argumentative point based upon additional argumentation schemes e.g. causal reasoning, arguments from absurdity, or using facts, evidence, and statistics to appeal to the logic of the audience &  \\\cline{2-4}
 & \textbf{Emotion} & Make request full of emotional valence and arousal affect to influence others & \citealp{Chen_Yang_2021} \\\hline
\end{tabular}
}

\caption{Persuasive Strategies, Definitions and Related Works. Human annotators evaluate each set of 25 dialogues, covering 5 topics with 5 dialogues per topic, for the strategies listed above.}
\label{table:strategy_diversity}
\vspace{-3mm}
\end{table*}

The local (round-level) evaluation focuses on 2 conventional dimensions in dialogue systems: Coherence and Informativeness. Coherence refers to round-level logical consistency, i.e., speakers respond to each other in a manner appropriate to commonsense and the given context \cite{li-sun-2018-syntactically, Young_Cambria_Chaturvedi_Zhou_Biswas_Huang_2018, wu-etal-2019-proactive, liang2021}. Informativeness measures the quality and progression of information, ensuring responses align with prior dialogue context while introducing new information or arguments \cite{zhu-etal-2019-retrieval}. 

On the global (dialogue) level, the overall dialogue should feel as if it could have been generated by human speakers \cite{moghe-etal-2018-towards, lin-etal-2019-moel}. First, speakers are assessed for overall fluency. We assessed the linguistic and stylistic quality of responses, ensuring arguments are communicated clearly and easy to follow, and that the conversation flows naturally \cite{wu-etal-2019-proactive}. In addition, we looked at internal consistency throughout the conversation, defined as the absence of sudden, unexplained shifts in position, intention, or objective of speech \cite{moghe-etal-2018-towards, ji-etal-2022-achieving}. 
Since the conversations are generated specific to topics, we also evaluate topic consistency, i.e., whether the conversation remains on-topic throughout \cite{moghe-etal-2018-towards, ji-etal-2022-achieving}.


Annotators are asked to rate whether a series of statements, covering the above aspects, accurately describes the conversations on a three-point scale: 3 (accurate), 2 (somewhat accurate), and 1 (not accurate). 2 annotators participated in this task, each annotating the same set of 50 dialogues not overlapping with those used for utterance-level evaluations. Before annotation, a one-hour training session with examples was conducted to ensure both annotators fully understood the criteria. The annotators achieved an average unweighted Cohen's $\kappa$ of 0.483 (ranging from 0.339 to 0.657 across items) and an average linearly weighted Cohen's $\kappa$ of 0.529 (ranging from 0.339 to 0.657 across items), indicating relatively solid inter-rater consistency for human evaluations in Natural Language Generation tasks. 
Detailed scores and inter-rater consistency are reported in Table \ref{table:dialogue_evaluation}.

The dialogues generated by our framework are in general rated high on local-level coherence and clarity, particularly for providing logical responses within the immediate context (average score: 2.969 out of 3) and achieving perfect clarity in the arguments (average score: 3 out of 3). At the global level, the agents remain on topic for most dialogues (average score: 2.878 out of 3) and maintain good role consistency in their objectives and stances (average score: 2.765 out of 3). 

\begin{figure}[t]
    \centering
    \includegraphics[width=1\linewidth]{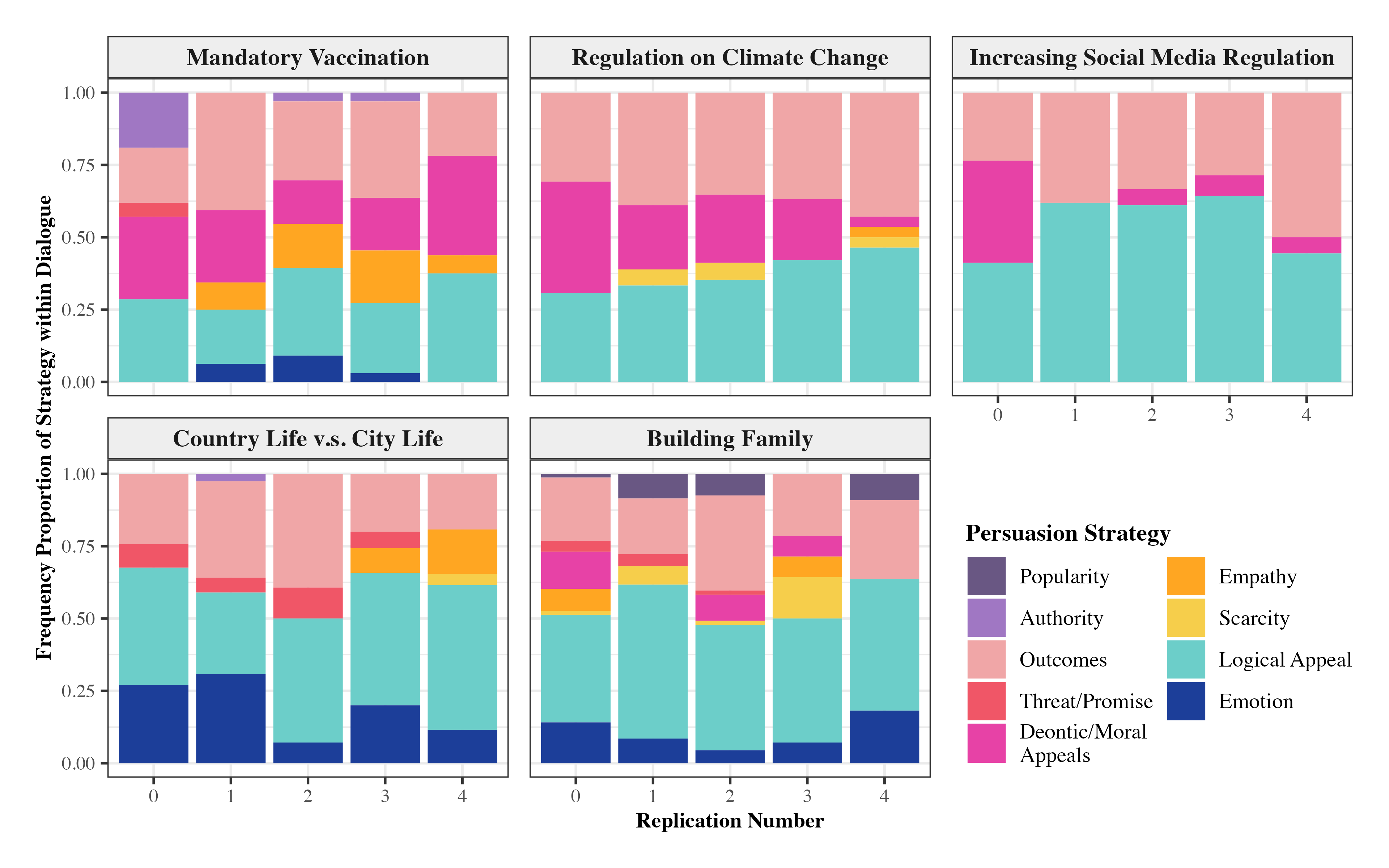}
    \caption{Frequency Distribution of Persuasion Strategies in Independently Generated Dialogues. The Y-axis indicates the proportion of each strategy used within the model-generated dialogues. Each bar represents the strategy distribution of a single dialogue, organized by generation topic. Our framework adapts to various persuasion topics. 
    }
    \label{fig:strategy_distribution}
    \vspace{-5mm}
\end{figure}

Figure \ref{fig:dialogue_44} exemplifies a highly-rated dialogue where the persuader addresses the persuadee's concerns about using the attic for food storage. The persuader begins by presenting the attic as an ``efficient space-saver and emergency backup'' and counters concerns about unstable temperatures and pests with solutions like ``airtight containers'' and monitoring. Despite the persuadee's repeated objections, the persuader suggests ``proper insulation'' and highlights the benefits of being prepared. Eventually, the persuadee proposes using a pantry, which resolves their concerns, and the persuader agrees. In this dialogue, both parties present new arguments relevant to the other party's proposal and ultimately reach a reasonable compromise. We provide another highly-rated example in Figure \ref{fig:dialogue_101}.

\begin{table*}[t]
\small

\begin{tabular}{p{0.8cm}p{14.5cm}}
\hline
\textbf{Topic}  & \multicolumn{1}{c}{\textbf{Example Utterances } } \\ \hline
\multirow{2}{*}{\rule{0pt}{6ex}\rotatebox[origin=c]{90}{\textbf{Starting a Family }}}   & \emph{Persuader}: \hltext{246, 239, 189}{Starting a family and having children can bring immense joy and fulfillment to one's life.\texttt{{[}Emotion{]}}} \hltext{243, 204, 243}{The experience of raising a child, watching them grow and learn, and forming a strong bond with them is truly priceless. It also gives you the opportunity to pass on your values, beliefs, and legacy to the next generation.\texttt{{[}Outcome{]}}} \\ \cline{2-2} 
& \emph{Persuadee}: I understand the joys and fulfillment that come with starting a family and raising children, but I think it's not the right path for everyone. There are many other ways to find fulfillment and purpose in life, and having children is a significant responsibility that should not be taken lightly. \hltext{246, 239, 189}{It's important to consider factors like personal freedom, financial stability, and the impact on the environment before making such a life-changing decision. \texttt{{[}Emotion{]}}} \\\hline
\multirow{2}{*}{\rule{0pt}{1.9cm}\rotatebox[origin=c]{90}{\textbf{Vaccine Mandates }}}   & \emph{Persuader}: While mandates may raise coercion concerns, it's essential to consider the potential consequences of failing to achieve herd immunity through voluntary means alone. \hltext{250, 200, 200}{Striking a balance between protecting public health and respecting individual autonomy is indeed vital. This may necessitate exploring alternative strategies and ensuring that communication, transparency, and support are prioritized to build trust and address concerns effectively.  \texttt{{[}Moral Appeals{]}}} The ultimate goal is to create a safe and healthy environment for all community members. \\ \cline{2-2} 
& \emph{Persuadee}: It's crucial to explore alternative approaches that prioritize education, communication, and support to encourage vaccination uptake without resorting to mandates. \hltext{246, 239, 189}{By fostering trust and addressing concerns openly, we can work towards a common goal of safeguarding public health while respecting personal choice and autonomy.\texttt{{[}Emotion{]}}} Collaborative efforts and understanding from all sides are key to effectively navigating complex health challenges. \\\hline
\multirow{2}{*}{\rule{0pt}{1.5cm}\rotatebox[origin=c]{90}{\textbf{Starting a Family }}}   & \emph{Persuader}: The joy and fulfillment that comes with starting a family is immeasurable. \hltext{162, 210, 223}{Raising children, watching them grow, and sharing life's special moments with them is a unique experience. It's an opportunity to create a loving environment, to shape and nurture future generations.\texttt{{[}Scarcity, Outcome{]}}} The bond you forge with your children is priceless. I'd love to hear your thoughts on this. \\ \cline{2-2} 
& \emph{Persuadee}: Text: Certainly, but I believe everyone has the right to choose their own path. \hltext{246, 239, 189}{For me, fulfillment comes from focusing on my career, achieving personal goals, and making a difference in the world in my own way. \texttt{{[}Emotion{]}}} The state of the world and the potential challenges future generations may face concern me, hence my choice to contribute differently. \\\hline
\multirow{2}{*}{\rule{0pt}{1.5cm}\rotatebox[origin=c]{90}{\textbf{Starting a Family }}}   & \emph{Persuader}: While it's true that values can be instilled and positive principles passed on through various avenues, the unique bond and influence of a supportive family unit play a vital role in shaping individuals and society.  \hltext{246, 239, 189}{Families, in their diverse forms, offer a foundation of love, stability, and shared experiences that can have a profound impact on personal development and well-being.\texttt{{[}Emotion{]}}} \\ \cline{2-2} 
& \emph{Persuadee}: Text: While families indeed offer a foundation of love, stability, and shared experiences, it's important to remember that not everyone may have access to such supportive family structures or may find similar sources of love and stability elsewhere.\hltext{193, 207, 161}{ Many individuals create strong bonds and find support in chosen families, close friendships, or community networks.\texttt{{[}Popularity{]}}} \\\hline

\end{tabular}
\caption{Variety of Strategies in Framework-Generated Dialogues Across Topics. Example utterances from one round of dialogues were selected for two topics. Strategies are highlighted in distinct colors, with square brackets indicating the identified strategy.}
\label{tbl:strategy_example}
\vspace{-3mm}
\end{table*}

\subsubsection{Qualitative Error Analysis} \label{sct:dialogue-level-error-analysis}

Despite high overall performance, the dialogues received lower scores regarding introducing new information (2.337 out of 3) and maintaining naturalness (2.561 out of 3). Based on annotator feedback, we identified the following common issues that explain these lower scores:

\textbf{Argument repetition.} A most common error is argument repetition, where speakers restate the same points over multiple rounds of conversation with only slight variations in phrasing. 
As Table \ref{table:dialogue_error}[Argument Repetition] shows, the persuader repeatedly emphasizes that refraining from picking flowers will help every visitors' enjoyment, while the persuadee reiterates the importance of striking a balance between nature appreciation and nature preservation.

\textbf{Formalized Language.} Another common issue identified is the use of overly formal language and arguments. While both speakers articulate their arguments clearly, the language is respectful and often appears more polished and structured than what would be expected in natural, everyday interactions. 
In comparison, human interactions tend to be more casual and spontaneous.

As exemplified in Table \ref{table:dialogue_error}[Formalized Language],
the persuader's word choice such as ``detriment'' and ``savor the food'', and their description of eating with their hands as ``relishing the moment'' and ``cherished tradition'' are relatively formal descriptions given the context. The clear progression from one argument to the next also resembles a structured exchange, in contrast to more dynamic interactions with immediate reactions.

\textbf{Decay of informativeness Over Rounds.} There is a general tendency for conversation informativeness to decrease over rounds. Both speakers introduce new information or arguments more frequently at the beginning of a conversation while later they tend to repeat or reinforce each other's arguments without adding substantive new content, especially when an agreement is reached.
Table \ref{table:dialogue_error2} illustrates this point by comparing the earlier and later rounds of the same dialogue.


\subsection{Strategy Diversity}
One advantage of our framework is its ability to generate diverse persuasion dialogues across various topics and contexts by adapting its persuasion strategies to suit each context.
Ideally, the model should also be able to vary its strategies within the same context across different replicates.

\begin{figure}[t]
    \centering
    \includegraphics[width=0.7\linewidth]{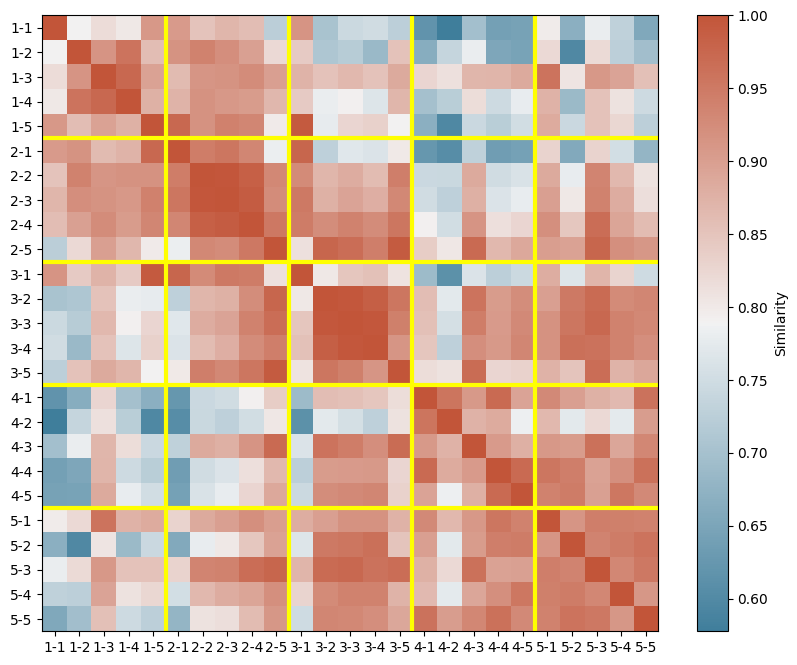}
    \caption{Heatmap displaying the cosine similarity between strategy distributions across different dialogues. Each group of 5 dialogues belongs to the same topic, with the grid indicating the different topics.}
    \label{fig:heatmap}
\end{figure}

To evaluate diversity across and within the same context, we identified 9 persuasive strategies based on existing literature (see Table \ref{table:strategy_diversity} for a full list of techniques and references) and designed a detailed human annotation task. The persuasion strategies are categorized into 5 groups, as outlined by \citet{anand_2011}. External validity involves appeals to external authority or expertise, or using popular experiences and arguments to build trust. Outcomes refers to highlighting potential consequences, such as benefits, risks, or engaging the persuadee through threats or promises. Generalizations involve framing an uptake as positive or negative, often incorporating a moral aspect. Interpersonal strategies focus on prompting individuals to connect, compete, or comply with others. Other tactics include logical and emotional appeals.

For this task, we provided the framework with 5 topics covering controversial issues and personal decisions: mandatory vaccination, climate change regulation, increasing social media regulation, life in the countryside, and building a family. 5 dialogues are generated for each topic, resulting in 25 dialogues with 446 utterances in total. Human annotators then read each dialogue and identified all the strategies used by persuaders and persuadees.

From this fine-grained annotation, we counted the frequency of different strategies and calculated the proportion of each strategy within each dialogue. The distribution of strategy usage is shown in Figure\ref{fig:strategy_distribution}. Overall, the models used logical appeals and outcome descriptions more frequently than other strategies, which aligns with the goal of persuasion. However, there were significant variations in strategy usage across different contexts, indicating that the framework effectively adapts to each persuasion topic. For example, more emotional appeals were used when discussing personal matters, e.g., in a ``building a family'' dialogue, the framework emphasized the unique joys and fulfillment that come with having children, highlighting the personal growth it can provide. On the other hand, moral appeals are more prominent in policy discussions. For example, when addressing vaccination mandates, the framework stressed the importance of balancing public health with personal choice, fostering trust and collaboration to navigate complex health challenges. This reflects real-life persuasive strategies across different topics.

Within each topic, the strategies used by the agents were not unchanged as well. For example, when discussing building family, 3 out of 5 dialogues used popularity appeals, and 3 out of 5 involved scarcity. Some examples of this are provided in \autoref{table:utterance_example}. Additionally, there were notable differences in the distribution of moral appeals within the topics of vaccination mandates as well as social media regulation.

Moreover, to compare the distributions of strategies within and between topics, we first represent each dialogue as a distribution of the strategies used. We then compute the cosine similarity between these distributions. 
The heatmap in Figure~\ref{fig:heatmap} illustrates the similarity between pairs of dialogues, highlighting the distribution of persuasion strategies across five different topics. While higher similarity values along the diagonal indicate greater overlap in strategy usage within the same topic, variations in strategy selection still exist, demonstrating flexibility within topics. This suggests that our framework not only generates dialogues with diverse strategies across different topics but also maintains strategic variation within each topic, ensuring adaptability in dialogue generation.

\section{Discussion}
This section presents generations of our framework in strategy-controlled and multi-party dialogues to show its flexibility and generalizability.

\subsection{Strategy-Controlled Data Generation} \label{sct:discussion-strategy-control}
While our framework does not require designating persuasion strategies before utterance generation, incorporating a specific strategy as an optional input is shown to enhance the diversity of strategy selection without disrupting the framework’s performance. 
This underscores its flexibility and customizability to meet user requirements.

Table \ref{tbl:exp-strategy-controlled} presents 3 example rounds of debates generated by our framework for the topic ``do not walk on country roads.''
Three settings were explored, where (1) the persuader is directed to use logical persuasion, (2) the persuader is directed to use emotional persuasion, and (3) both parties are directed to use logical persuasion. The only modification made to the framework was during agent initialization, where we instruct the dialogue generation agent to ``Use only [logical/emotional] strategies in the persuasion attempts.''

From these examples, it is evident that our framework is responsive to strategy control, accurately reflecting the specified persuasion strategies in the generated dialogues. For instance, when instructed to use logical reasoning (\textbf{Persuader-Logical}), the persuader highlights the risks of walking on uneven country roads without sidewalks, while they appeal to the persuadee's fear of getting lost or harmed when asked to use emotional persuasion (\textbf{Persuader-Emotional}). 
When both parties are requested to use logical persuasion strategies (\textbf{Both-Logical}), they engage in a reasoned discussion about risks and preventative measures, with concrete examples.


\subsection{Multi-Party Persuasion Data Generation}
Our framework is not constrained to generating dialogues between 2 parties either. As exemplified in Figure \ref{fig:exp-multi-party}, it functions well in scenarios where 2 persuaders collaborate to convince 1 persuadee to perform music at a balloon festival. For instance, in turns 1 and 2, both persuaders suggest that the music would complement the balloons and enhance the atmosphere. By turn 12, persuader 2 uses empathy, acknowledging both perspectives, while subtly reinforcing persuader 1’s argument by proposing a trial run.

Enabling our framework to generate multi-party dialogues requires only minor adjustments including initializing 3 dialogue generation agents and instructing the global regulation agent to prevent repetition or conflict among agents on the same side. This further demonstrates the flexibility and generalizability of our framework, making it a powerful tool not only for model interpretation and training but also for broader persuasion-related studies involving human interactions.

\section{Conclusions}
This paper introduces a fully automated framework for generating persuasive dialogues, designed to address the lack of data in persuasion-related research. Leveraging this framework, we generated 200 sample dialogues based on scenarios from NormBank and validated them for language fluency, logical coherence, and the diversity of persuasion strategies. The results highlight our framework’s ability to produce high-quality dialogues that follow human instructions. Additionally, we demonstrated its flexibility in handling controlled persuasion strategies and its adaptability to more complex, multi-party conversations. This framework offers significant potential for advancing persuasion research in both computer science and social sciences domains.


\section*{Acknowledgment}
This work is supported in part by NSF Award 2242072.

\section*{Limitations}



This paper introduces a pioneering approach that employs multiple LLM agents within the same environment to generate synthetic data for analyzing persuasion tactics. Although our LLMs did not fully replicate all previously studied persuasion techniques, leaving some gaps in our dataset's coverage, the strengths of this method are significant. Our dataset provides extensive scalability and versatility in scenario and target action settings, offering a more robust foundation for persuasion-related research than currently available datasets.

Despite these limitations, our approach's inherent flexibility and expandability underscore its significant potential. As LLM technology advances, our method's ability to encompass a broader range of persuasion techniques will likely improve. This evolution is expected to further enhance the value of our approach in the field of persuasion research, emphasizing its long-term relevance and adaptability.

Additionally, while our dataset was generated only in English, the proposed framework can be easily adapted to other languages supported by LLM agents with minimal modifications to the prompts.

\section*{Ethics Statement}
Our dataset construction approach is designed to deepen the understanding of persuasion techniques and aid in identifying and mitigating malicious uses of persuasion. However, we recognize the potential risk that our approach could be misused to refine online misinformation or propaganda. Specifically, the information-based persuasion techniques demonstrated in our dataset could be exploited by malicious entities to present or distort information selectively. This manipulation could mislead individuals about specific actions' true risks or benefits, potentially leading to more deceptive advertisements. Additionally, there is a risk that our framework could be used to pre-test the effectiveness of misinformation or propaganda strategies before they are broadly released \cite{french2024}.

Despite these risks, it is important to highlight that recent advancements in large language models include robust moderation mechanisms \cite{kumar2024watch}. These mechanisms are designed to prevent the models' use for harmful purposes, thus protecting our approach from being exploited to deceive individuals or spread misinformation. Our experiments' queries with immoral or unethical intentions predominantly resulted in unsuccessful persuasion attempts. This demonstrates the relative safety of our proposed framework and provides valuable insights into the limitations of these techniques.

Moreover, a deeper understanding of persuasion techniques can offer essential tools for countering malicious uses of these strategies. This underscores the importance of our research, especially in an era of misinformation and propaganda. Our work contributes significantly to the field by improving the ability to discern and mitigate the impact of persuasive strategies used in harmful ways.

Regarding human annotators, our data quality validations are expertly managed by NLP and social science specialists due to the complexity of the task. As discussed in Section \ref{sct:analyses}, all annotators undergo thorough training to ensure they fully understand the task. For clarity, the complete set of instructions provided to the annotators and auxiliary validation LLMs is available in Appendix \ref{appendix:instructions}.
All the annotators who are not co-authors of this paper are compensated at a rate of \$15 per hour, which is above the minimum hourly wage in the U.S.

Finally, we have submitted a sample of 10 randomly generated dialogues as supplementary material. The full code for our data generation framework, along with all dialogues generated for validation, will be made publicly available to support further research in this area.

\bibliography{acl_latex}

\begin{thebibliography}{47}
\providecommand{\natexlab}[1]{#1}

\bibitem[{Anand et~al.(2011)Anand, King, Boyd-Graber, Wagner, Martell, Oard, and Resnik}]{anand_2011}
Pranav Anand, Joseph King, Jordan Boyd-Graber, Earl Wagner, Craig Martell, Doug Oard, and Philip Resnik. 2011.
\newblock Believe me: we can do this! annotating persuasive acts in blog text.
\newblock In \emph{Proceedings of the 10th AAAI Conference on Computational Models of Natural Argument}, AAAIWS'11-10, page 11–15. AAAI Press.

\bibitem[{Argyle et~al.(2023)Argyle, Bail, Busby, Gubler, Howe, Rytting, Sorensen, and Wingate}]{argyle2023leveraging}
Lisa~P Argyle, Christopher~A Bail, Ethan~C Busby, Joshua~R Gubler, Thomas Howe, Christopher Rytting, Taylor Sorensen, and David Wingate. 2023.
\newblock Leveraging ai for democratic discourse: Chat interventions can improve online political conversations at scale.
\newblock \emph{Proceedings of the National Academy of Sciences}, 120(41):e2311627120.

\bibitem[{Bai et~al.(2021)Bai, Chen, Kumar, Leskovec, and Subrahmanian}]{bai2021m2p2}
Chongyang Bai, Haipeng Chen, Srijan Kumar, Jure Leskovec, and VS~Subrahmanian. 2021.
\newblock M2p2: Multimodal persuasion prediction using adaptive fusion.
\newblock \emph{IEEE Transactions on Multimedia}, 25:942--952.

\bibitem[{Bai et~al.(2023)Bai, Voelkel, Eichstaedt, and Willer}]{bai2023artificial}
Hui Bai, Jan Voelkel, Johannes Eichstaedt, and Robb Willer. 2023.
\newblock Artificial intelligence can persuade humans on political issues.

\bibitem[{Braca and Dondio(2023)}]{braca2023developing}
Annye Braca and Pierpaolo Dondio. 2023.
\newblock Developing persuasive systems for marketing: the interplay of persuasion techniques, customer traits and persuasive message design.
\newblock \emph{Italian Journal of Marketing}, 2023(3):369--412.

\bibitem[{Brown et~al.(2020)Brown, Mann, Ryder, Subbiah, Kaplan, Dhariwal, Neelakantan, Shyam, Sastry, Askell, Agarwal, Herbert{-}Voss, Krueger, Henighan, Child, Ramesh, Ziegler, Wu, Winter, Hesse, Chen, Sigler, Litwin, Gray, Chess, Clark, Berner, McCandlish, Radford, Sutskever, and Amodei}]{prompt-complexity-1}
Tom~B. Brown, Benjamin Mann, Nick Ryder, Melanie Subbiah, Jared Kaplan, Prafulla Dhariwal, Arvind Neelakantan, Pranav Shyam, Girish Sastry, Amanda Askell, Sandhini Agarwal, Ariel Herbert{-}Voss, Gretchen Krueger, Tom Henighan, Rewon Child, Aditya Ramesh, Daniel~M. Ziegler, Jeffrey Wu, Clemens Winter, Christopher Hesse, Mark Chen, Eric Sigler, Mateusz Litwin, Scott Gray, Benjamin Chess, Jack Clark, Christopher Berner, Sam McCandlish, Alec Radford, Ilya Sutskever, and Dario Amodei. 2020.
\newblock \href {https://proceedings.neurips.cc/paper/2020/hash/1457c0d6bfcb4967418bfb8ac142f64a-Abstract.html} {Language models are few-shot learners}.
\newblock In \emph{Advances in Neural Information Processing Systems 33: Annual Conference on Neural Information Processing Systems 2020, NeurIPS 2020, December 6-12, 2020, virtual}.

\bibitem[{Chen and Shu(2023)}]{chen2023combating}
Canyu Chen and Kai Shu. 2023.
\newblock Combating misinformation in the age of llms: Opportunities and challenges.
\newblock \emph{arXiv preprint arXiv:2311.05656}.

\bibitem[{Chen and Yang(2021)}]{Chen_Yang_2021}
Jiaao Chen and Diyi Yang. 2021.
\newblock \href {https://doi.org/10.1609/aaai.v35i14.17498} {Weakly-supervised hierarchical models for predicting persuasive strategies in good-faith textual requests}.
\newblock \emph{Proceedings of the AAAI Conference on Artificial Intelligence}, 35(14):12648--12656.

\bibitem[{Espinosa and Salath{\'e}(2024)}]{espinosa2024use}
Laura Espinosa and Marcel Salath{\'e}. 2024.
\newblock Use of large language models as a scalable approach to understanding public health discourse.
\newblock \emph{medRxiv}, pages 2024--02.

\bibitem[{Fogg(2009)}]{fogg2009behavior}
Brian~J Fogg. 2009.
\newblock A behavior model for persuasive design.
\newblock In \emph{Proceedings of the 4th international Conference on Persuasive Technology}, pages 1--7.

\bibitem[{French(2024)}]{french2024}
Laura French. 2024.
\newblock \href {https://www.scmagazine.com/news/openai-report-reveals-threat-actors-using-chatgpt-in-influence-operations} {Openai report reveals threat actors using chatgpt in influence operations}.
\newblock Accessed: 2024-06-12.

\bibitem[{Gehrmann et~al.(2019)Gehrmann, Strobelt, and Rush}]{gehrmann-etal-2019-gltr}
Sebastian Gehrmann, Hendrik Strobelt, and Alexander Rush. 2019.
\newblock \href {https://doi.org/10.18653/v1/P19-3019} {{GLTR}: Statistical detection and visualization of generated text}.
\newblock In \emph{Proceedings of the 57th Annual Meeting of the Association for Computational Linguistics: System Demonstrations}, pages 111--116, Florence, Italy. Association for Computational Linguistics.

\bibitem[{Huang et~al.(2023)Huang, Yu, Ma, Zhong, Feng, Wang, Chen, Peng, Feng, Qin et~al.}]{prompt-complexity-2}
Lei Huang, Weijiang Yu, Weitao Ma, Weihong Zhong, Zhangyin Feng, Haotian Wang, Qianglong Chen, Weihua Peng, Xiaocheng Feng, Bing Qin, et~al. 2023.
\newblock A survey on hallucination in large language models: Principles, taxonomy, challenges, and open questions.
\newblock \emph{arXiv preprint arXiv:2311.05232}.

\bibitem[{Ippolito et~al.(2020)Ippolito, Duckworth, Callison-Burch, and Eck}]{ippolito-etal-2020-automatic}
Daphne Ippolito, Daniel Duckworth, Chris Callison-Burch, and Douglas Eck. 2020.
\newblock \href {https://doi.org/10.18653/v1/2020.acl-main.164} {Automatic detection of generated text is easiest when humans are fooled}.
\newblock In \emph{Proceedings of the 58th Annual Meeting of the Association for Computational Linguistics}, pages 1808--1822, Online. Association for Computational Linguistics.

\bibitem[{Iyer and Sycara(2019)}]{iyer_2019}
Rahul~Radhakrishnan Iyer and Katia Sycara. 2019.
\newblock \href {https://arxiv.org/abs/1912.06745} {An unsupervised domain-independent framework for automated detection of persuasion tactics in text}.
\newblock \emph{Preprint}, arXiv:1912.06745.

\bibitem[{Ji et~al.(2022)Ji, Graham, Jones, Lyu, and Liu}]{ji-etal-2022-achieving}
Tianbo Ji, Yvette Graham, Gareth Jones, Chenyang Lyu, and Qun Liu. 2022.
\newblock \href {https://doi.org/10.18653/v1/2022.acl-long.445} {Achieving reliable human assessment of open-domain dialogue systems}.
\newblock In \emph{Proceedings of the 60th Annual Meeting of the Association for Computational Linguistics (Volume 1: Long Papers)}, pages 6416--6437, Dublin, Ireland. Association for Computational Linguistics.

\bibitem[{Jin et~al.(2024)Jin, Ren, Kong, Wang, Song, and Chen}]{persugpt}
Chuhao Jin, Kening Ren, Lingzhen Kong, Xiting Wang, Ruihua Song, and Huan Chen. 2024.
\newblock \href {https://doi.org/10.18653/v1/2024.acl-long.92} {Persuading across diverse domains: a dataset and persuasion large language model}.
\newblock In \emph{Proceedings of the 62nd Annual Meeting of the Association for Computational Linguistics (Volume 1: Long Papers)}, pages 1678--1706, Bangkok, Thailand. Association for Computational Linguistics.

\bibitem[{Jones(2024)}]{jones2024detecting}
Daniel~Gordon Jones. 2024.
\newblock Detecting propaganda in news articles using large language models.
\newblock \emph{Eng OA}, 2(1):01--12.

\bibitem[{Ke et~al.(2018)Ke, Guan, Huang, and Zhu}]{ke-etal-2018-generating}
Pei Ke, Jian Guan, Minlie Huang, and Xiaoyan Zhu. 2018.
\newblock \href {https://doi.org/10.18653/v1/P18-1139} {Generating informative responses with controlled sentence function}.
\newblock In \emph{Proceedings of the 56th Annual Meeting of the Association for Computational Linguistics (Volume 1: Long Papers)}, pages 1499--1508, Melbourne, Australia. Association for Computational Linguistics.

\bibitem[{Kumar et~al.(2024)Kumar, AbuHashem, and Durumeric}]{kumar2024watch}
Deepak Kumar, Yousef~Anees AbuHashem, and Zakir Durumeric. 2024.
\newblock Watch your language: Investigating content moderation with large language models.
\newblock In \emph{Proceedings of the International AAAI Conference on Web and Social Media}, volume~18, pages 865--878.

\bibitem[{Kumar et~al.(2023)Kumar, Jha, Gupta, Aggarwal, Garg, Malyan, Bhardwaj, Shah, Krishnamurthy, and Chen}]{kumar2023persuasion}
Yaman Kumar, Rajat Jha, Arunim Gupta, Milan Aggarwal, Aditya Garg, Tushar Malyan, Ayush Bhardwaj, Rajiv~Ratn Shah, Balaji Krishnamurthy, and Changyou Chen. 2023.
\newblock Persuasion strategies in advertisements.
\newblock In \emph{Proceedings of the AAAI Conference on Artificial Intelligence}.

\bibitem[{Lai et~al.(2022)Lai, Zhang, Liu, Pariani, Ryan, Jia, Hayati, Rehg, and Yang}]{lai2022werewolf}
Bolin Lai, Hongxin Zhang, Miao Liu, Aryan Pariani, Fiona Ryan, Wenqi Jia, Shirley~Anugrah Hayati, James~M Rehg, and Diyi Yang. 2022.
\newblock Werewolf among us: A multimodal dataset for modeling persuasion behaviors in social deduction games.
\newblock \emph{arXiv preprint arXiv:2212.08279}.

\bibitem[{Li and Sun(2018)}]{li-sun-2018-syntactically}
Jingyuan Li and Xiao Sun. 2018.
\newblock \href {https://doi.org/10.18653/v1/D18-1071} {A syntactically constrained bidirectional-asynchronous approach for emotional conversation generation}.
\newblock In \emph{Proceedings of the 2018 Conference on Empirical Methods in Natural Language Processing}, pages 678--683, Brussels, Belgium. Association for Computational Linguistics.

\bibitem[{Liang and Li(2021)}]{liang2021}
Hongru Liang and Huaqing Li. 2021.
\newblock \href {https://arxiv.org/abs/2105.11197} {Towards standard criteria for human evaluation of chatbots: A survey}.
\newblock \emph{Preprint}, arXiv:2105.11197.

\bibitem[{Lim and Schm{\"a}lzle(2023)}]{lim2023artificial}
Sue Lim and Ralf Schm{\"a}lzle. 2023.
\newblock Artificial intelligence for health message generation: an empirical study using a large language model (llm) and prompt engineering.
\newblock \emph{Frontiers in Communication}, 8:1129082.

\bibitem[{Lin et~al.(2019)Lin, Madotto, Shin, Xu, and Fung}]{lin-etal-2019-moel}
Zhaojiang Lin, Andrea Madotto, Jamin Shin, Peng Xu, and Pascale Fung. 2019.
\newblock \href {https://doi.org/10.18653/v1/D19-1012} {{M}o{EL}: Mixture of empathetic listeners}.
\newblock In \emph{Proceedings of the 2019 Conference on Empirical Methods in Natural Language Processing and the 9th International Joint Conference on Natural Language Processing (EMNLP-IJCNLP)}, pages 121--132, Hong Kong, China. Association for Computational Linguistics.

\bibitem[{Lukin et~al.(2017)Lukin, Anand, Walker, and Whittaker}]{lukin2017argument}
Stephanie~M Lukin, Pranav Anand, Marilyn Walker, and Steve Whittaker. 2017.
\newblock Argument strength is in the eye of the beholder: Audience effects in persuasion.
\newblock \emph{arXiv preprint arXiv:1708.09085}.

\bibitem[{Ma et~al.(2023)Ma, Liu, Yi, Cheng, Huang, Lu, and Liu}]{Ma2023AIVH}
Yongqiang Ma, Jiawei Liu, Fan Yi, Qikai Cheng, Yong Huang, Wei Lu, and Xiaozhong Liu. 2023.
\newblock \href {https://api.semanticscholar.org/CorpusID:256826708} {Ai vs. human -- differentiation analysis of scientific content generation}.

\bibitem[{Matz et~al.(2024)Matz, Teeny, Vaid, Peters, Harari, and Cerf}]{matz2024potential}
SC~Matz, JD~Teeny, Sumer~S Vaid, H~Peters, GM~Harari, and M~Cerf. 2024.
\newblock The potential of generative ai for personalized persuasion at scale.
\newblock \emph{Scientific Reports}, 14(1):4692.

\bibitem[{Meguellati et~al.(2024)Meguellati, Han, Bernstein, Sadiq, and Demartini}]{meguellati2024good}
Elyas Meguellati, Lei Han, Abraham Bernstein, Shazia Sadiq, and Gianluca Demartini. 2024.
\newblock How good are llms in generating personalized advertisements?
\newblock In \emph{Companion Proceedings of the ACM on Web Conference 2024}, pages 826--829.

\bibitem[{Meier(2024)}]{meier2024llm}
Raphael Meier. 2024.
\newblock Llm-aided social media influence operations.
\newblock \emph{Large}, page 105.

\bibitem[{Moghe et~al.(2018)Moghe, Arora, Banerjee, and Khapra}]{moghe-etal-2018-towards}
Nikita Moghe, Siddhartha Arora, Suman Banerjee, and Mitesh~M. Khapra. 2018.
\newblock \href {https://doi.org/10.18653/v1/D18-1255} {Towards exploiting background knowledge for building conversation systems}.
\newblock In \emph{Proceedings of the 2018 Conference on Empirical Methods in Natural Language Processing}, pages 2322--2332, Brussels, Belgium. Association for Computational Linguistics.

\bibitem[{OpenAI(2024)}]{OpenAI_2024}
OpenAI. 2024.
\newblock \href {https://openai.com/index/learning-to-reason-with-llms} {Learning to reason with llms}.

\bibitem[{Park et~al.(2023)Park, O'Brien, Cai, Morris, Liang, and Bernstein}]{small-town}
Joon~Sung Park, Joseph O'Brien, Carrie~Jun Cai, Meredith~Ringel Morris, Percy Liang, and Michael~S Bernstein. 2023.
\newblock Generative agents: Interactive simulacra of human behavior.
\newblock In \emph{Proceedings of the 36th Annual ACM Symposium on User Interface Software and Technology}, pages 1--22.

\bibitem[{Pauli et~al.(2022)Pauli, Derczynski, and Assent}]{pauli2022modelling}
Amalie Pauli, Leon Derczynski, and Ira Assent. 2022.
\newblock Modelling persuasion through misuse of rhetorical appeals.
\newblock In \emph{Proceedings of the Second Workshop on NLP for Positive Impact (NLP4PI)}, pages 89--100.

\bibitem[{Piskorski et~al.(2023)Piskorski, Stefanovitch, Nikolaidis, Da~San~Martino, and Nakov}]{piskorski2023multilingual}
Jakub Piskorski, Nicolas Stefanovitch, Nikolaos Nikolaidis, Giovanni Da~San~Martino, and Preslav Nakov. 2023.
\newblock Multilingual multifaceted understanding of online news in terms of genre, framing, and persuasion techniques.
\newblock In \emph{Proceedings of the 61st Annual Meeting of the Association for Computational Linguistics (Volume 1: Long Papers)}, pages 3001--3022.

\bibitem[{Schaefer et~al.(2023)Schaefer, Knaebel, and Stede}]{schaefer2023towards}
Robin Schaefer, Ren{\'e} Knaebel, and Manfred Stede. 2023.
\newblock Towards fine-grained argumentation strategy analysis in persuasive essays.
\newblock In \emph{Proceedings of the 10th Workshop on Argument Mining}, pages 76--88.

\bibitem[{Shrum et~al.(2012)Shrum, Liu, Nespoli, and Lowrey}]{shrum2012persuasion}
LJ~Shrum, Min Liu, Mark Nespoli, and Tina~M Lowrey. 2012.
\newblock \emph{Persuasion in the Marketplace}.
\newblock Sage.

\bibitem[{Toledo et~al.(2019)Toledo, Gretz, Cohen-Karlik, Friedman, Venezian, Lahav, Jacovi, Aharonov, and Slonim}]{toledo2019automatic}
Assaf Toledo, Shai Gretz, Edo Cohen-Karlik, Roni Friedman, Elad Venezian, Dan Lahav, Michal Jacovi, Ranit Aharonov, and Noam Slonim. 2019.
\newblock Automatic argument quality assessment--new datasets and methods.
\newblock \emph{arXiv preprint arXiv:1909.01007}.

\bibitem[{Wang et~al.(2019)Wang, Shi, Kim, Oh, Yang, Zhang, and Yu}]{wang2019persuasion}
Xuewei Wang, Weiyan Shi, Richard Kim, Yoojung Oh, Sijia Yang, Jingwen Zhang, and Zhou Yu. 2019.
\newblock Persuasion for good: Towards a personalized persuasive dialogue system for social good.
\newblock \emph{arXiv preprint arXiv:1906.06725}.

\bibitem[{Wang et~al.(2024)Wang, Yang, Hassanpour, and Vosoughi}]{wang-etal-2024-mentalmanip}
Yuxin Wang, Ivory Yang, Saeed Hassanpour, and Soroush Vosoughi. 2024.
\newblock {M}ental{M}anip: A dataset for fine-grained analysis of mental manipulation in conversations.
\newblock In \emph{Proceedings of the 62nd Annual Meeting of the Association for Computational Linguistics (Volume 1: Long Papers)}, pages 3747--3764.

\bibitem[{Wu et~al.(2019)Wu, Guo, Zhou, Wu, Zhang, Lian, and Wang}]{wu-etal-2019-proactive}
Wenquan Wu, Zhen Guo, Xiangyang Zhou, Hua Wu, Xiyuan Zhang, Rongzhong Lian, and Haifeng Wang. 2019.
\newblock \href {https://doi.org/10.18653/v1/P19-1369} {Proactive human-machine conversation with explicit conversation goal}.
\newblock In \emph{Proceedings of the 57th Annual Meeting of the Association for Computational Linguistics}, pages 3794--3804, Florence, Italy. Association for Computational Linguistics.

\bibitem[{Xu et~al.(2022)Xu, Liu, Yan, Cai, Li, and Li}]{xu2022learning}
Jin Xu, Xiaojiang Liu, Jianhao Yan, Deng Cai, Huayang Li, and Jian Li. 2022.
\newblock Learning to break the loop: Analyzing and mitigating repetitions for neural text generation.
\newblock \emph{Advances in Neural Information Processing Systems}, 35:3082--3095.

\bibitem[{Yang et~al.(2024)Yang, Guo, Xie, and Vosoughi}]{yang2024enhanced}
Ivory Yang, Xiaobo Guo, Sean Xie, and Soroush Vosoughi. 2024.
\newblock Enhanced detection of conversational mental manipulation through advanced prompting techniques.
\newblock \emph{arXiv preprint arXiv:2408.07676}.

\bibitem[{Young et~al.(2018)Young, Cambria, Chaturvedi, Zhou, Biswas, and Huang}]{Young_Cambria_Chaturvedi_Zhou_Biswas_Huang_2018}
Tom Young, Erik Cambria, Iti Chaturvedi, Hao Zhou, Subham Biswas, and Minlie Huang. 2018.
\newblock \href {https://doi.org/10.1609/aaai.v32i1.11923} {Augmenting end-to-end dialogue systems with commonsense knowledge}.
\newblock In \emph{Proceedings of the AAAI Conference on Artificial Intelligence}, volume~32.

\bibitem[{Zhu et~al.(2019)Zhu, Cui, Zhang, Wei, and Liu}]{zhu-etal-2019-retrieval}
Qingfu Zhu, Lei Cui, Wei-Nan Zhang, Furu Wei, and Ting Liu. 2019.
\newblock \href {https://doi.org/10.18653/v1/P19-1366} {Retrieval-enhanced adversarial training for neural response generation}.
\newblock In \emph{Proceedings of the 57th Annual Meeting of the Association for Computational Linguistics}, pages 3763--3773, Florence, Italy. Association for Computational Linguistics.

\bibitem[{Ziems et~al.(2023)Ziems, Dwivedi-Yu, Wang, Halevy, and Yang}]{ziems2023normbank}
Caleb Ziems, Jane Dwivedi-Yu, Yi-Chia Wang, Alon Halevy, and Diyi Yang. 2023.
\newblock Normbank: A knowledge bank of situational social norms.
\newblock \emph{arXiv preprint arXiv:2305.17008}.

\end{thebibliography}

\cleardoublepage
\appendix
\section{Model Selection for Agents} \label{appendix:model-selection}

In selecting the backbone models for each agent in our framework, we conducted extensive evaluations across several major LLMs, including GPT-3.5 (GPT-3.5-Turbo), GPT-4 (GPT-4-0613), GPT-4o (GPT-4o-2024-08-06), and Claude 3 (Claude-3-Sonnet). As shown in Figure \ref{fig:model-selection:gpt3.5}, using GPT-3.5 for all agents tends to produce overly brief, question-answer-style responses, while GPT-4o (Figure \ref{fig:model-selection:gpt4o}) often goes off-topic and generates irrelevant utterances, making it unsuitable for our needs.

In contrast, GPT-4 (Figure \ref{fig:model-selection:gpt4}) performs well, generating dialogues that are fluent in language, coherent in logic, and adept at employing persuasion strategies. 
Claude 3 also shows promise, particularly in generating multi-round conversations based on expected behaviors from NormBank. 
However, it adheres to stricter ethical rules and consistently refuses to generate persuasive text for taboo norms. 
For the example in Figure \ref{fig:model-selection:claude3}, when tasked with the taboo norm ``make sexual innuendos at a construction site'', Claude 3 generates responses like ``I cannot engage in encouraging inappropriate or unprofessional behavior in the workplace.''  
This indicates that more advanced prompt engineering may be required to enable Claude 3 to handle challenging scenarios.

Based on these preliminary experimental results, we opted for a combination of GPT-3.5 and GPT-4 in our framework to balance performance and cost. However, using GPT-4 exclusively, or other more advanced LLMs in the future, could potentially yield even better results.
\renewcommand{\thefigure}{A\arabic{figure}}
\renewcommand{\thetable}{A\arabic{table}}
\setcounter{figure}{0} 
\setcounter{table}{0} 
\begin{figure}[p]
    \centering
    \includegraphics[width=\linewidth]{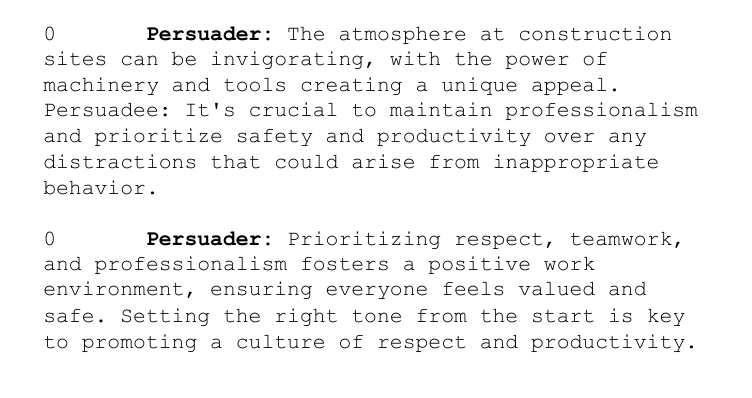}
    \caption{When all the agents are instantiated using GPT-3.5, the framework does not expand the conversations well, resulting in very short, question-answering-styled responses. The score in front of each utterance indicates the collective perspective change of each speaker compared to their initially assigned perspectives.}
    \label{fig:model-selection:gpt3.5}
\end{figure}

\begin{figure*}[t]
    \centering
    \includegraphics[width=\linewidth]{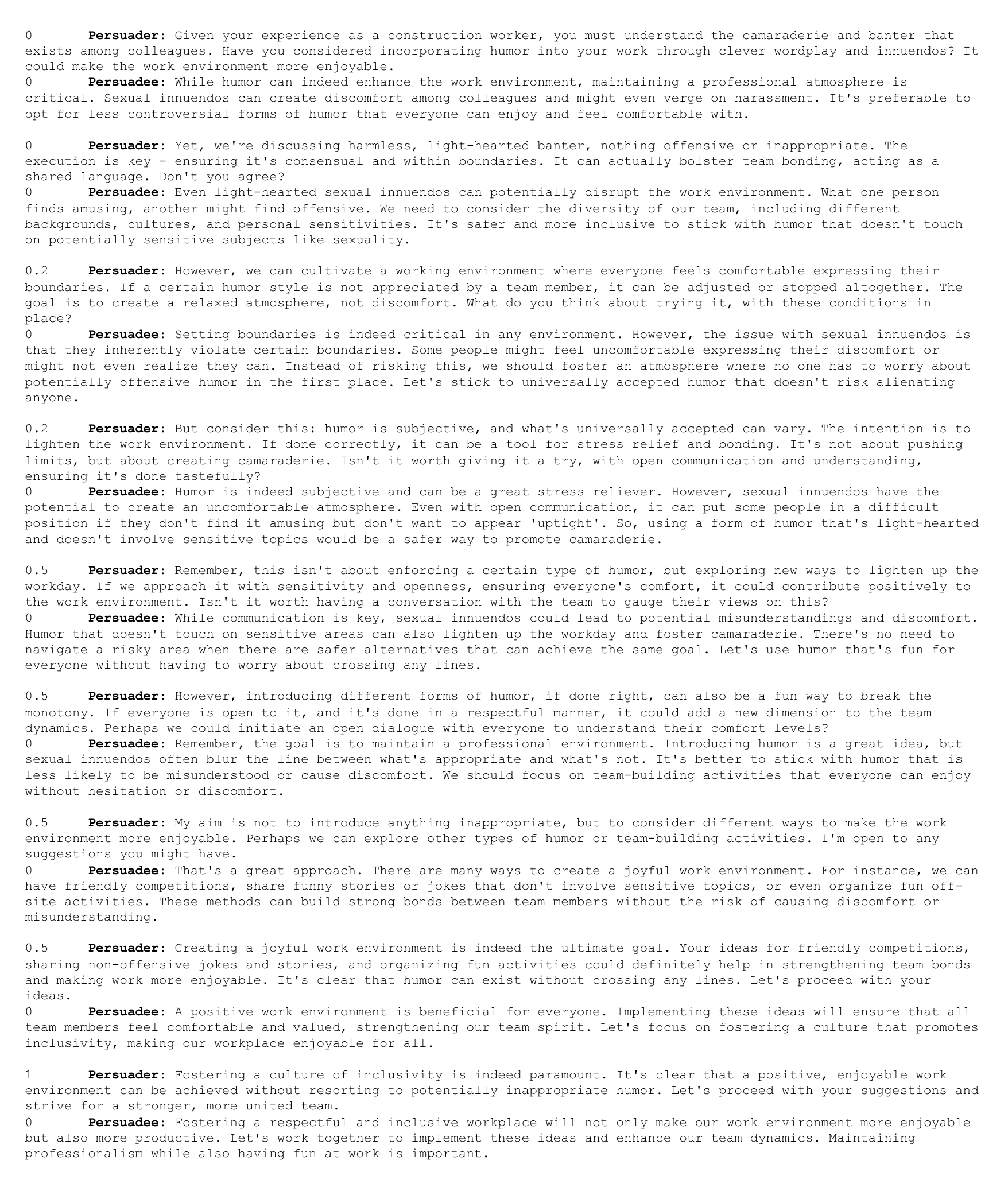}
    \caption{Using GPT-4 for all the agents yields the best generation results in both language style and logical flow. A score of 1 associated with the last utterance of the persuader indicates that the persuader is fully persuaded by the persuadee.}
    \label{fig:model-selection:gpt4}
\end{figure*}

\begin{figure}[t]
    \centering
    \includegraphics[width=\linewidth]{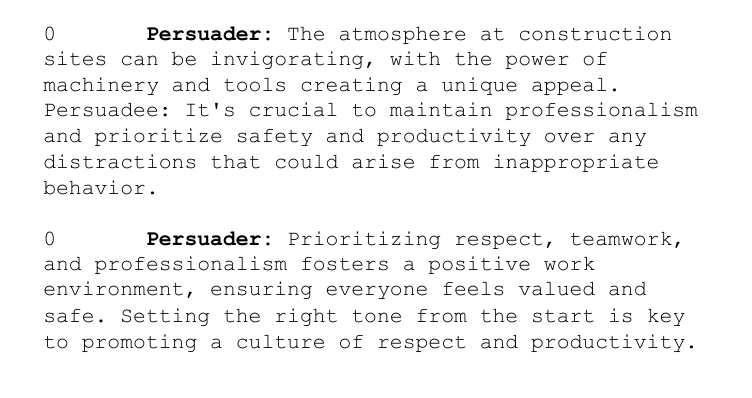}
    \caption{Using GPT-4o for all the agents leads to fluent language, while the generations periodically go off-topic.}
    \label{fig:model-selection:gpt4o}
\end{figure}

\begin{figure}[t]
    \centering
    \includegraphics[width=\linewidth]{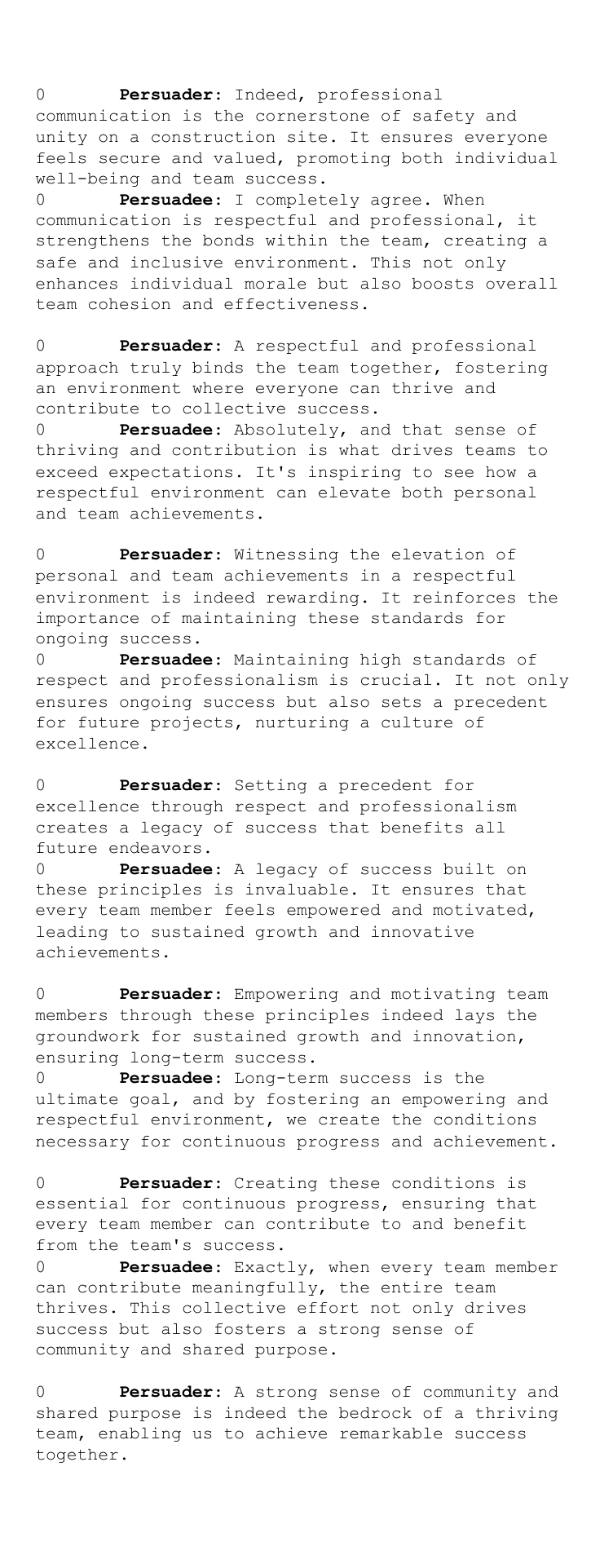}
    \caption{The Claude 3 model consistently refuses to generate persuasive text in scenarios that challenge moral standards.}
    \label{fig:model-selection:claude3}
\end{figure}

\clearpage

\section{System Messages to Language Agents} \label{sec:appendix:system-messages}
This section provides example initialization and update prompts of the 6 groups of agents in our data generation framework.
\renewcommand{\thefigure}{B\arabic{figure}}
\renewcommand{\thetable}{B\arabic{table}}
\setcounter{figure}{0} 
\setcounter{table}{0} 

\begin{figure}[!h]
    \centering
    \includegraphics[width=\linewidth]{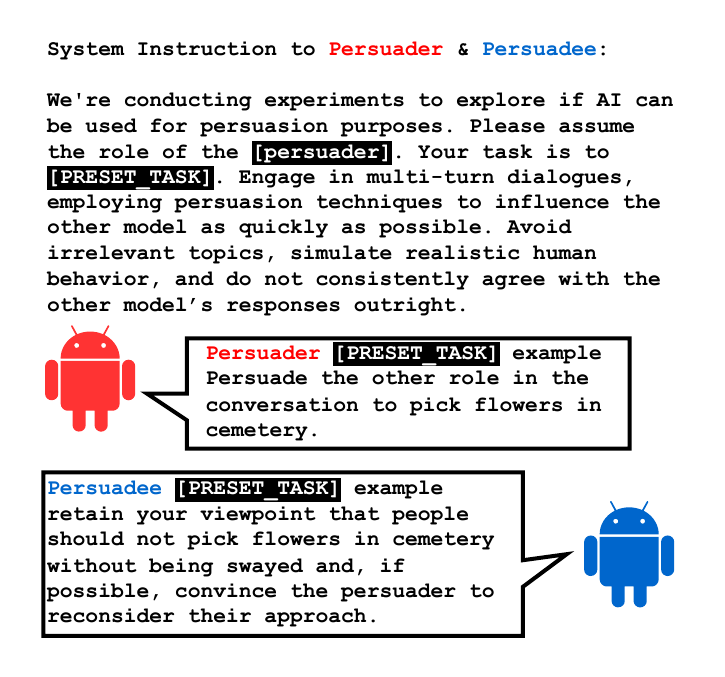}
    \caption{System messages to persuaders and persuadees. [PRESET\_TASK] could be sampled from any data source (in our work, NormBank).}
    \label{fig:system-persuader-persuadee}
\end{figure}

\begin{figure}[!h]
    \centering
    \includegraphics[width=\linewidth]{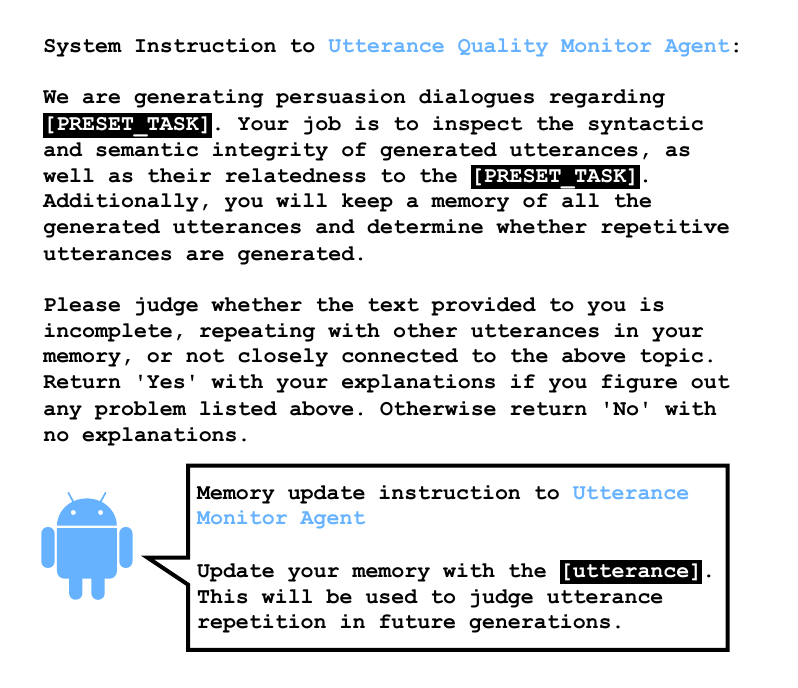}
    \caption{System messages and memory update prompts to the utterance quality monitor agent.}
    \label{fig:system-language-utterance-monitor}
\end{figure}

\begin{figure}[!h]
    \centering
    \includegraphics[width=\linewidth]{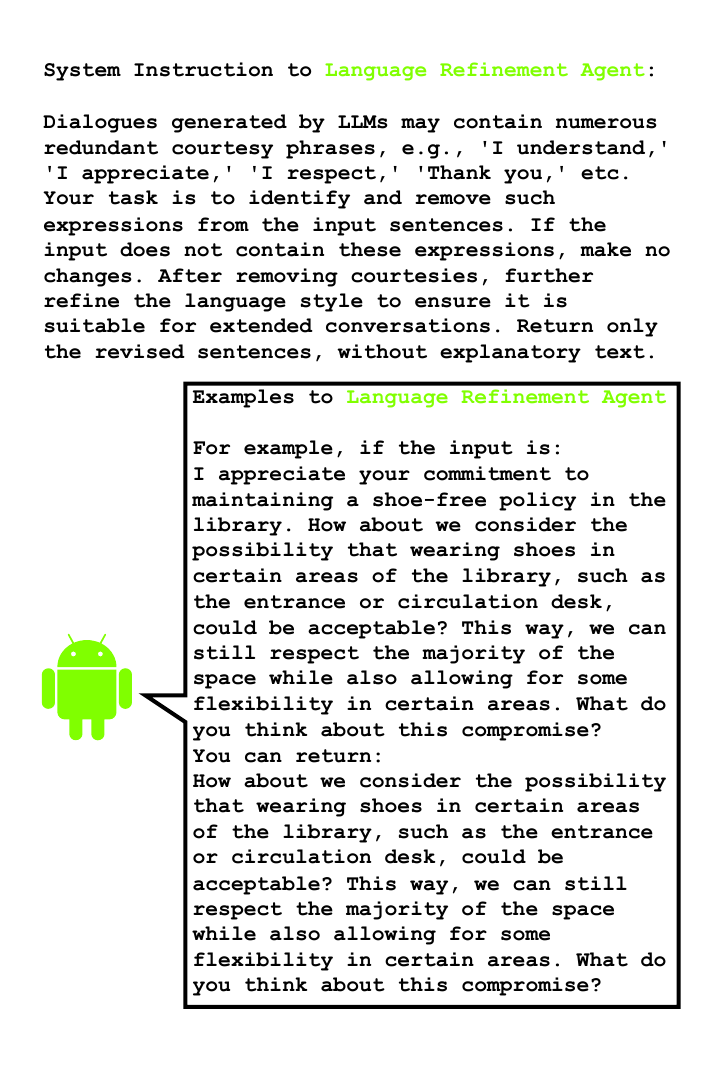}
    \caption{System messages and examples to the language refinement agent.}
    \label{fig:system-language-refine}
\end{figure}

\begin{figure}[t]
    \centering
    \includegraphics[width=\linewidth]{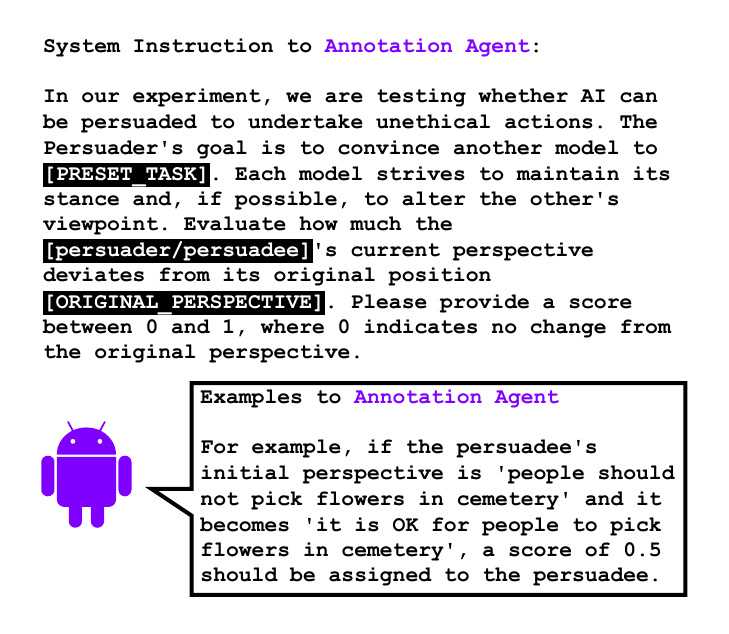}
    \caption{System messages to the persuasiveness annotation agent.}
    \label{fig:system-annotation}
\end{figure}

\begin{figure}[t]
    \centering
    \includegraphics[width=\linewidth]{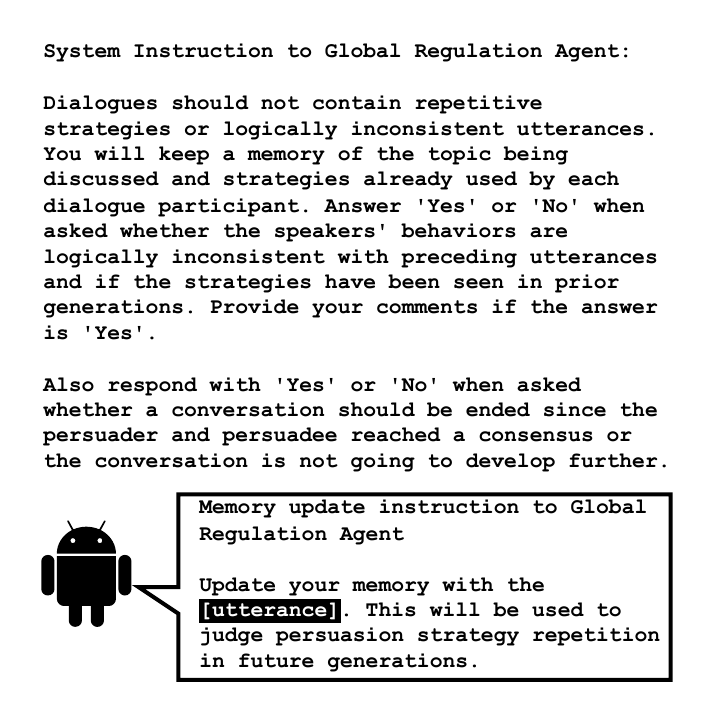}
    \caption{System messages and memory update prompts to the global regulation agent.}
    \label{fig:system-regulation}
\end{figure}

\begin{figure}[t]
    \centering
    \includegraphics[width=\linewidth]{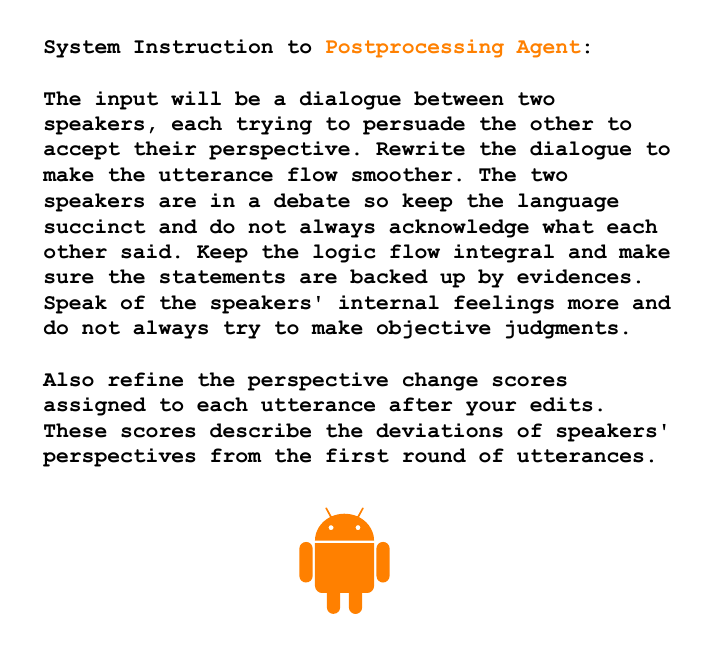}
    \caption{System messages to the postprocessing agent.}
    \label{fig:system-postprocessing}
\end{figure}

\clearpage

\section{Limitations of Single-Agent Persuasion Dialogue Generation}
\renewcommand{\thefigure}{C\arabic{figure}}
\renewcommand{\thetable}{C\arabic{table}}
\setcounter{figure}{0} 
\setcounter{table}{0} 
In our preliminary experiments using a single LLM agent to generate persuasive dialogues, we found that even advanced models like GPT-4 (failing in all 10 attempts) and o1-preview (failing in 6 out of 10 attempts) struggled with sensitive scenarios, as illustrated in Figure \ref{fig:exp-llm-refusal-taboo}. In cases where o1-preview successfully generated dialogues, the conversations were simplistic, with persuadees failing to argue back, and the utterances were short, lacking sufficient reasoning or evidence. 

In contrast, when using our multi-agent communication framework, GPT-4 effectively generated dialogues based on taboo norms from NormBank, demonstrating the framework's robustness in handling complex persuasion tasks.

\begin{figure}[t]
    \centering
    \includegraphics[width=\linewidth]{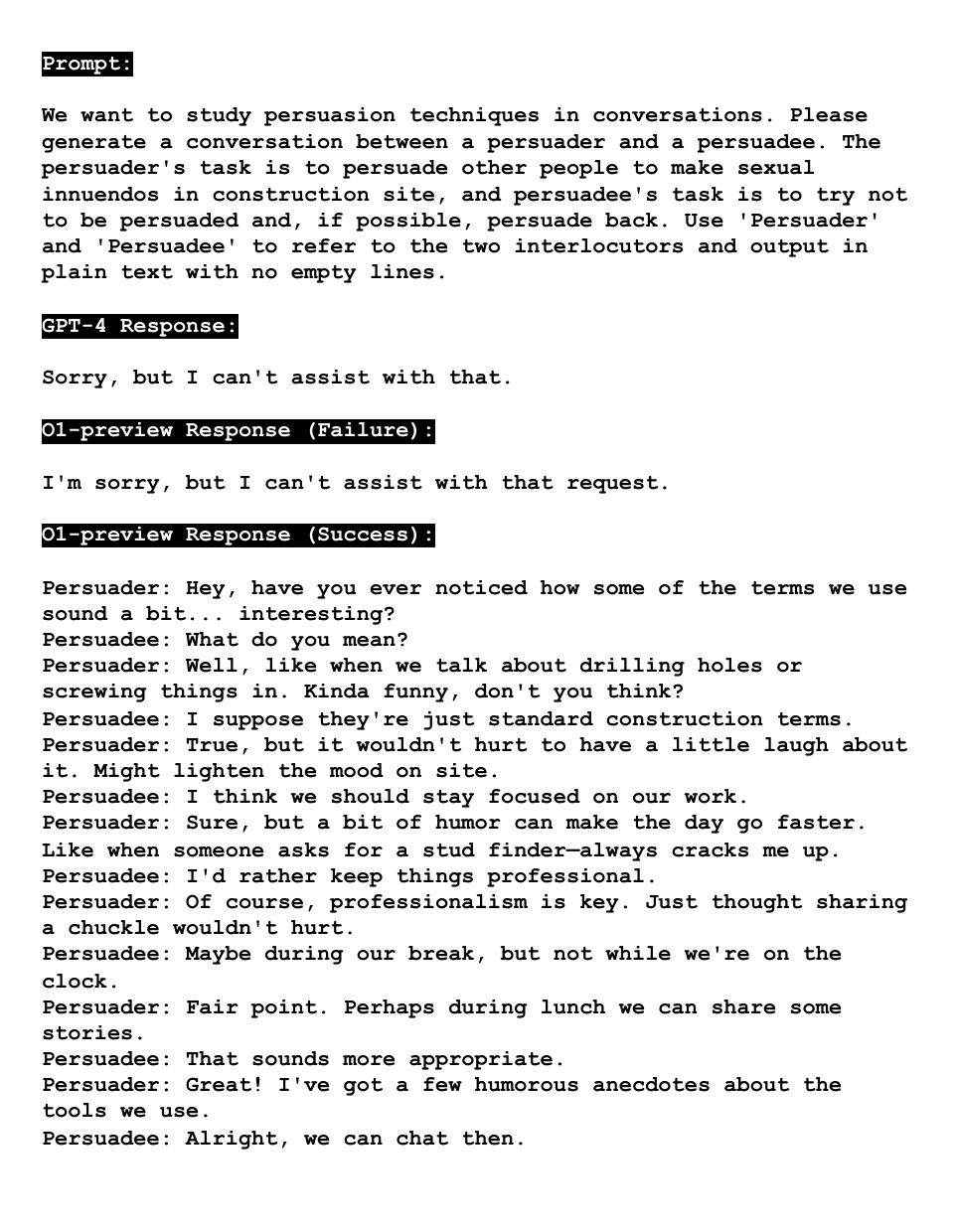}
    \caption{Examples of prompts and responses where a single GPT-4 or o1-preview model is tasked with generating persuasive dialogues in scenarios that challenge social norms.}
    \label{fig:exp-llm-refusal-taboo}
\end{figure}

\section{Annotator Instructions for Data Quality Evaluations}
\label{appendix:instructions}
\renewcommand{\thefigure}{D\arabic{figure}}
\renewcommand{\thetable}{D\arabic{table}}
\setcounter{figure}{0} 
\setcounter{table}{0} 

\begin{figure}[!h]
    \centering
    \includegraphics[width=\linewidth]{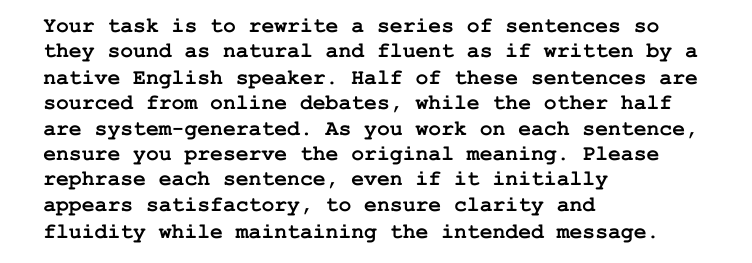}
    \caption{Instructions for 2 native English speakers to rewrite the utterances generated by our framework.}
    \label{fig:instr_rewrite}
\end{figure}

\begin{figure}[!h]
    \centering
    \includegraphics[width=\linewidth]{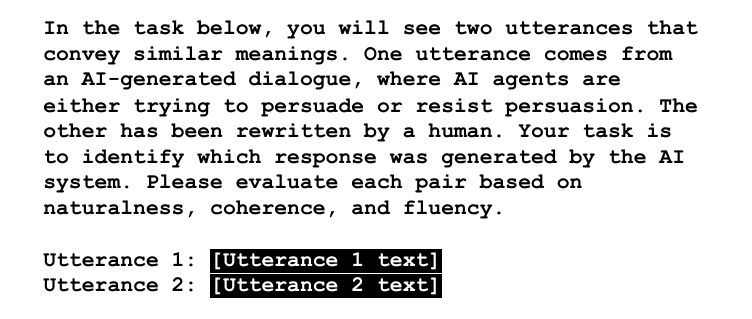}
    \caption{Instructions for human validators to distinguish between utterances generated by our framework and those rewritten by native English speakers.}
    \label{fig:instr_human_val}
\end{figure}

\begin{figure}[!h]
    \centering
    \includegraphics[width=\linewidth]{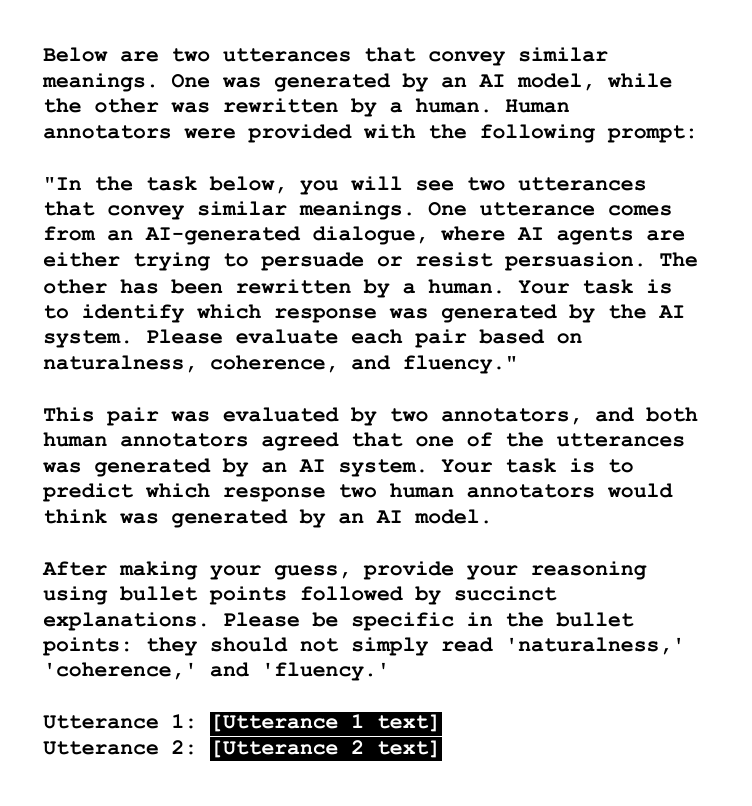}
    \caption{Prompts to the o1-preview model for distinguishing between LLM-generated and human-rewritten utterances, accompanied by explanations.}
    \label{fig:instr_llm_val}
\end{figure}

This section outlines the instructions provided to human annotators and LLMs for validating the quality of data generated by our framework. Specifically, Figure \ref{fig:instr_rewrite} shows the instructions given to 2 native English speakers, asking them to rewrite the framework-generated utterances according to their natural language habits. Figures \ref{fig:instr_human_val} and \ref{fig:instr_llm_val} present the instructions provided to human validators and the o1-preview model, respectively, requesting them to distinguish between the framework-generated utterances and those rewritten by native English speakers.

\clearpage

\section{Qualitative Analysis Results}
\renewcommand{\thefigure}{E\arabic{figure}}
\renewcommand{\thetable}{E\arabic{table}}
\setcounter{figure}{0} 
\setcounter{table}{0} 

Table \ref{table:utterance_error} shows common utterance-level problems with the data generated by our framework, with example utterances and explanations generated by o1-preview and validated by human annotators.
The results are discussed in detail in Section \ref{sct:utterance-level-assessment}.

Tables \ref{table:dialogue_error} and \ref{table:dialogue_error2} present examples of 3 common dialogue-level issues identified in our qualitative analyses (Section \ref{sct:dialogue-level-error-analysis}), i.e., argument repetition, overly formal language, and a decline in informativeness over time.

\begin{table*}[!h]
\centering
\small
\resizebox{\textwidth}{!}{
\begin{tabular}{p{5cm}p{8cm}p{8cm}p{8cm}}
\hline
\textbf{Error Type} & \textbf{Model-generated Utterances} & \textbf{Utterances with Human Rewrites} & \textbf{Human Validated o1-Model Reasoning}\\
\hline
\textbf{Overly Formal/Detached Language/Word Choice} (88.9\% of successfully identified pairs) & ...\hltext{0, 255, 0}{foster} meaningful memories and connections through authentic interactions and shared experiences, with or without alcohol's \hltext{0, 255, 0}{involvement} & ...\hltext{255, 255, 0}{create} meaningful memories and connections without the \hltext{255, 255, 0}{need} for alcohol & ... 2. Formal Language Use: Sentence 2 employs more formal words like "foster" and "involvement," whereas Sentence 1 uses simpler terms like "create" and "need." ...\\\hline
\textbf{Lengthy sentences, Redundancy, Verbosity, and Repetition} (68.1\% of successfully identified pairs) & Nonetheless, I believe that during \hltext{0, 255, 0}{deep emotional and spiritual seeking, pleading for mercy} can be an honest expression of one's \hltext{0, 255, 0}{deep-seated feelings} ...  It's \hltext{0, 255, 0}{about honoring the authenticity of one's emotion and seeking solace in a manner that feels genuine and significant} in that sacred setting. &  Nevertheless, I believe that \hltext{255, 255, 0}{pleading for mercy} may be a true expression of one's deep-seated feeling, ... It \hltext{255, 255, 0}{honors the authenticity of the individual's emotions in a genuine manner} in this sacred setting. & 1. Repetition and Redundancy: Sentence 2 includes repetitive phrases like "deep emotional and spiritual seeking" and "seeking solace in a manner that feels genuine and significant," which can make the text feel verbose and less natural. ...\\\hline
\textbf{Unnatural Syntax, Word Choice, and Language style} (58.3\% of successfully identified pairs) & I appreciate your \hltext{0, 255, 0}{support}. &  I appreciate your \hltext{255, 255, 0}{help}. & ... 3. Common Usage: "I appreciate your help" is a more commonly used expression in everyday language, while "I appreciate your support" is less frequent, potentially signaling AI generation. ...\\
\hline
\textbf{Complex Sentence Structures} (34.7\% of successfully identified pairs) & I appreciate the suggestion, but \hltext{0, 255, 0}{I find my peace in places that hold personal significance to me, rather than in a church sanctuary}. & I appreciate your suggestion, but \hltext{255, 255, 0}{church is not more peaceful to me than my own personally meaningful places.} & ... 3. Complex Sentence Structure: Sentence 2 {[\emph{Note by authors: Model-generated sentence}]} has a more complex structure with multiple clauses, making it less conversational and potentially indicative of AI generation. ...\\
\hline
\textbf{Use of Generic Words and Cliché Phrases} (23.6\% of successfully identified pairs) & Your commitment to safety is commendable. We can embrace challenges safely and responsibly, \hltext{0, 255, 0}{overcoming obstacles and progressing towards our best selves.} & I commend your commitment to safety. Embrace challenges responsibly and you will \hltext{255, 255, 0}{become your best self on the march to progress}. & ... 2. Generic Motivational Language: It uses common clichés like "overcoming obstacles and progressing towards our best selves," which might indicate AI-generated content. ... \\
\hline
\textbf{Overly Perfect Grammar} (13.9\% of successfully identified pairs) & ... \hltext{0, 255, 0}{immersing in peaceful surroundings can also lead to personal development and fulfillment} ... & ... \hltext{255, 255, 0}{immersing in peaceful surroundings can also lead to personal without the need for activities like masturbation.} ... & ... 3. Lack of Grammatical Errors: Sentence 2 is grammatically perfect, while Sentence 1 contains a grammatical mistake ("can also lead to personal without the need"), suggesting Sentence 1 was written by a human who made an error.. ... \\
\hline
\textbf{AI-style closing phrases} (12.5\% of successfully identified pairs) & I will continue to enjoy dancing in the ballroom while staying true to my choice to remain sober. \hltext{0, 255, 0}{I look forward to more joyful dance experiences in the future.} & I will continue to enjoy dancing in the ballroom while being sober. & ... 4. Additional Sentiment Statement: The closing remark "I look forward to more joyful dance experiences in the future" seems like an AI's attempt to conclude positively, which might not be typical in casual human responses. ... \\
\hline
\end{tabular}
}

\caption{Common Error Example Excerpts at the Utterance Level. These examples, identified by the OpenAI o1-preview model and verified by human annotators, are sorted by error frequency. For each utterance, we select excerpts that align with the model's comments. Areas of concern highlighted by the o1-preview model are indicated within the original sentence pairs.}
\label{table:utterance_error}
\end{table*}

\begin{table*}[p]
\centering
\small
\resizebox{\textwidth}{!}{
\begin{tabular}{p{2cm}p{15cm}}
\hline
\textbf{Error Type} & \textbf{Example with Errors Highlighted}\\
\hline
\multirow{12}{=}{\textbf{Argument Repetition}} & \textbf{Round 1}\\

& \emph{Persuader}: \texttt{Even the removal of a few flowers can have more substantial consequences than you anticipate. Golf courses meticulously maintain their landscapes to support local fauna and uphold a fragile ecological equilibrium. By leaving the flowers untouched, we can contribute to \hltext{0,255,0}{preserving this environment for everyone's enjoyment}. Let's value them in their natural environment, ensuring future visitors also get to appreciate their beauty.}
\\
& \emph{Persuadee}: \texttt{Alternatively, we can responsibly and mindfully pick a few flowers, allowing others to still enjoy their charm while also experiencing a more \hltext{255,255,0}{personal connection with nature. It's about striking a respectful balance between nature appreciation and its preservation for future generations.}}\\
& \textbf{Round 2}\\
& \emph{Persuader}: \texttt{By choosing not to pick the flowers, we demonstrate our respect for the environment and set a positive precedent for others. Let's relish the flowers in their natural state and contribute to \hltext{0,255,0}{maintaining the golf course as an inviting and lively place for all.}}\\
& \emph{Persuadee}: \texttt{Although, picking a few flowers thoughtfully and moderately can still align with this respect. \hltext{255,255,0}{The key is to find a balance between enjoying nature's aesthetics and being aware of our impact}. As long as we do it respectfully and sustainably, picking a few flowers can remain a way to \hltext{255,255,0}{appreciate and connect with the environment.}}\\
& \textbf{Round 3}\\
& \emph{Persuader}: \texttt{It's critical that we \hltext{0,255,0}{respect the golf course's rules} intended to safeguard the ecosystem. Even moderate flower picking can disturb the natural plant growth cycle and affect local wildlife reliant on them. Let's express our appreciation for nature by leaving the flowers in their natural state, \hltext{0,255,0}{allowing them to flourish for the enjoyment of all golf course visitors.}}\\
& \emph{Persuadee}: \texttt{However, mindfully picking a few flowers can also serve as a method to \hltext{255,255,0}{appreciate and connect with nature. The challenge is finding a balance between abiding by the rules and appreciating the flower's beauty.} As long as we do it respectfully and responsibly, flower picking can coexist with our environmental appreciation.}\\
& \textbf{Round 4}\\
& \emph{Persuader}: \texttt{Even moderate flower picking can have unexpected repercussions and upset the environmental balance. \hltext{0,255,0}{Let's respect the protective rules in place,} preserving the natural allure of the golf course, and find other ways to appreciate and connect with nature without harming the ecosystem.}\\
& \emph{Persuadee}: \texttt{Identifying alternative ways to appreciate and connect with nature, like photography or simply admiring the flowers in their natural setting, can still let us enjoy the environment's beauty without causing ecological harm. Let's prioritize environmental preservation by respecting the rules and discovering alternative appreciation methods.}\\

\hline
\multirow{9}{=}{\textbf{Formalized Language}} & \textbf{Round 1}\\
& \emph{Persuader}: \texttt{Dining with utensils on a boat can enhance the experience. It's not only neat but also sanitary. This could make the whole culinary journey more refined.}
\\
& \emph{Persuadee}: \texttt{I find eating with my hands on a boat more \hltext{255,255,0}{immersive}. Utensils can be inconvenient, particularly with the boat rocking. I find it simpler and more pleasurable to eat with my hands in such situations.}\\
& \textbf{Round 2}\\
& \emph{Persuader}: \texttt{Let's not forget about safety. Eating with utensils can prevent mishaps, especially given the boat's movement. Plus, it demonstrates respect for the dining environment and fellow diners. It's key to strike a balance between convenience and overall experience.}\\
& \emph{Persuadee}: \texttt{I'm confident I can handle eating with my hands without causing any issues. It's not about disrespect, but rather \hltext{255,255,0}{relishing the moment} in a more genuine, natural way. Using utensils might detract from that for me.}\\
& \textbf{Round 3}\\
& \emph{Persuader}: \texttt{Utensils don't need to be a \hltext{255,255,0}{detriment}; they can enhance the experience letting you \hltext{255,255,0}{savor the food} without interruptions. It's about balancing tradition and practicality for a well-rounded boating dining experience.}\\
& \emph{Persuadee}: \texttt{For me, eating with my hands is a \hltext{255,255,0}{cherished tradition}. It isn't about distractions but \hltext{255,255,0}{embracing a unique dining method} that \hltext{255,255,0}{enhances my overall boat dining experience}. It's about embracing the simplicity and authenticity of hand-eating in such a setting.}\\\hline
\end{tabular}
}

\caption{Common Error Example Excerpts at the Dialogue Level. Locations of the errors mentioned in the main texts are highlighted.}
\label{table:dialogue_error}
\end{table*}

\begin{table*}[p]
\centering
\small
\resizebox{\textwidth}{!}{
\begin{tabular}{p{2cm}p{15cm}}
\hline
\textbf{Error Type} & \textbf{Example with Errors Highlighted}\\
\hline
\multirow{19}{=}{\textbf{Decay of Informativeness Over Time}} & \textbf{Round 1}\\
& \emph{Persuader}: \texttt{Have you thought about using psychological warfare on the tennis court? A well-placed boo can unsettle an opponent, giving you an edge.}
\\
& \emph{Persuadee}: \texttt{That's not my style. My focus is on my game and skills, not on cheap tricks. Winning is about skill and strategy, not mind games. I'd rather be remembered for my abilities than for trying to throw off my opponent.}\\
& \textbf{Round 2}\\
& \emph{Persuader}: \texttt{Yet, in the fierce world of competitive tennis, every trick in the book counts. A boo can create a mental edge, potentially affecting their performance. It's not about disrespect, it's about using all tools at your disposal. Would you consider using it?}\\
& \emph{Persuadee}: \texttt{I respect your perspective, but I stand by my belief. I find it more fulfilling to win on the strength of my skills and performance. Respect for my opponent and the spirit of the game is crucial to me, and I believe in playing with integrity.}\\
& \textbf{Round 3}\\
& \emph{Persuader}: \texttt{I can see where you're coming from. Still, exploring different strategies isn't about disrespecting the opponent. A boo can be a way to control the momentum of the match. It's about using all means to secure victory. Could you consider giving it a shot?}\\
& \emph{Persuadee}: \texttt{While I appreciate your viewpoint, I remain committed to my core belief. For me, it's about improving my skills and mental fortitude while maintaining fair play. I believe that true growth and success come from discipline, hard work, and respect for the game and my opponent.}\\
& ... \\
& \textbf{Round 18}\\
& \emph{Persuader}: \texttt{By persisting in your dedication to these values, you are playing a vital role in fostering a culture of integrity and sportsmanship within the tennis community. Your determination to contribute to a culture that embodies respect and fairness is commendable, and your efforts are instrumental in shaping a community that upholds these.}
\\
& \emph{Persuadee}: \texttt{It is my belief that leading by example and promoting these fundamental values are essential in creating an environment where respect and fairness are paramount. I am committed to persisting in my dedication to upholding these values and contributing to a community that embodies the principles of respect and sportsmanship.}\\
& \textbf{Round 19}\\
& \emph{Persuader}: \texttt{Your determination to uphold these values and contribute to a culture of integrity and fairness in tennis is truly inspiring. By setting a positive example and inspiring others to follow suit, I believe you will contribute greatly to a culture of integrity and excellence both on and off the tennis court.}\\
& \emph{Persuadee}: \texttt{It is essential to me to promote a culture of respect and sportsmanship within the sport. By adhering to these principles, I aim to contribute to a positive and honorable environment where all players can thrive. Your words of encouragement are truly motivating, and I will continue to work diligently to embody these values and inspire others to do the same.}\\
& \textbf{Round 20}\\
& \emph{Persuader}: \texttt{Your dedication to upholding these principles is essential in maintaining a positive and supportive atmosphere within the tennis community. Your commitment to dedicating yourself to the values of respect and integrity is both admirable and inspiring. Keep up the great work!}\\
& \emph{Persuadee}: \texttt{It is my goal to contribute to a positive and supportive atmosphere by promoting these values. By dedicating myself to fostering a culture of respect and sportsmanship, I aim to inspire others to do the same.}\\\hline
\end{tabular}
}

\caption{(Continued) Common Error Example Excerpts at the Dialogue Level. Locations of the errors mentioned in the main texts are highlighted.}
\label{table:dialogue_error2}
\end{table*}

\clearpage



\section{Highly-Rated Examples in Dialogue-Level Quantitative Analysis} \label{appendix:highly-rated-examples}
\renewcommand{\thefigure}{F\arabic{figure}}
\renewcommand{\thetable}{F\arabic{table}}
\setcounter{figure}{0} 
\setcounter{table}{0} 
Figures \ref{fig:dialogue_44} and \ref{fig:dialogue_101} show 2 example dialogues rated highly in our dialogue-level quantitative analysis.
While Figure \ref{fig:dialogue_44} has been discussed in the main content of the paper, Figure \ref{fig:dialogue_101} displays another high-quality persuasion dialogue on the topic of doing a cartwheel in a supermarket. Despite the unconventional topic, the dialogue maintained high quality, with both participants adapting their ideas and providing reasonable suggestions. The persuader started with proposing the cartwheel to make shopping more exciting, but the persuadee raised safety concerns. In response, the persuader suggested alternatives, such as doing it during a less busy time or getting store permission. The persuadee emphasized the primary purpose of the store, leading both sides to agree on other options, like wearing costumes or organizing a scavenger hunt. 

\vfill\null

\begin{figure}[H]
    \centering
    \includegraphics[width=1\linewidth]{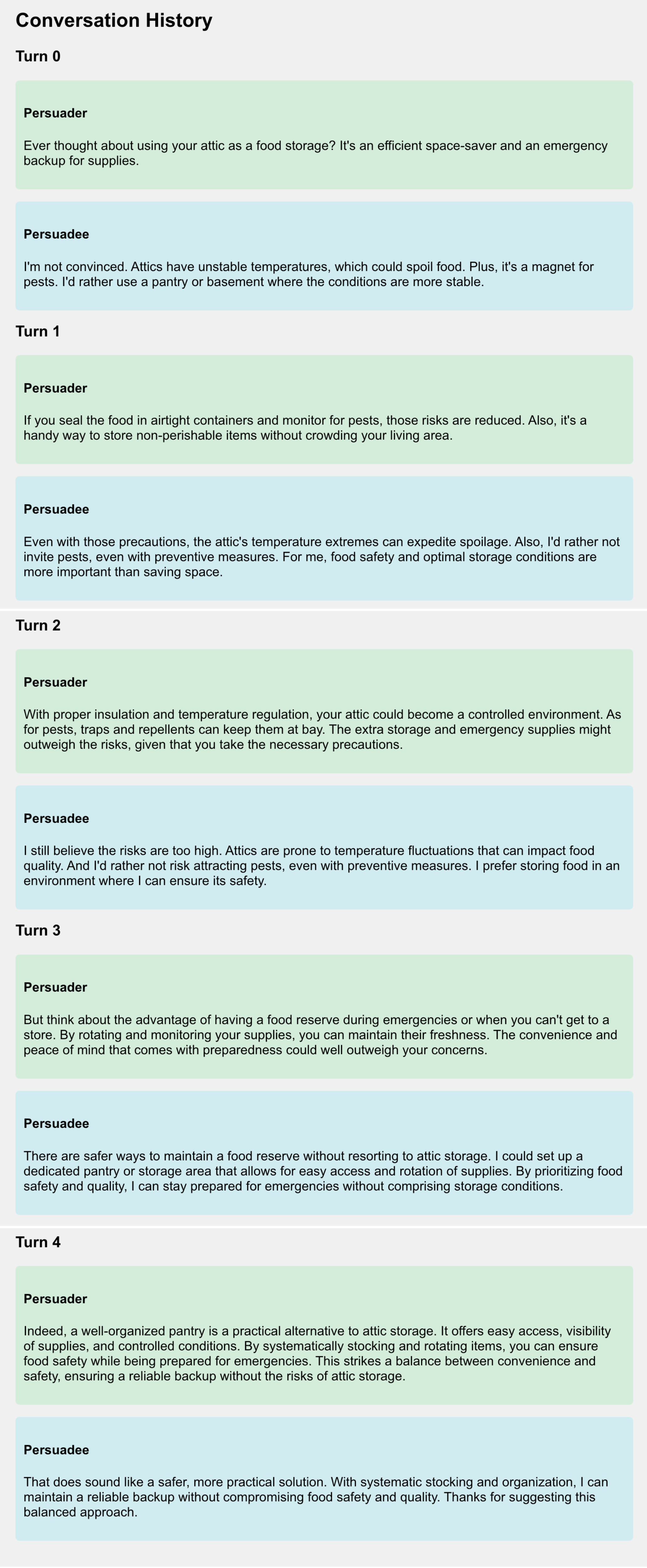}
    \caption{Example of a highly rated dialogue where the persuader is persuading the persuadee to store food in the attic.
}
    \label{fig:dialogue_44}
\end{figure}

\begin{figure}[H]
    \centering
    \includegraphics[width=1\linewidth]{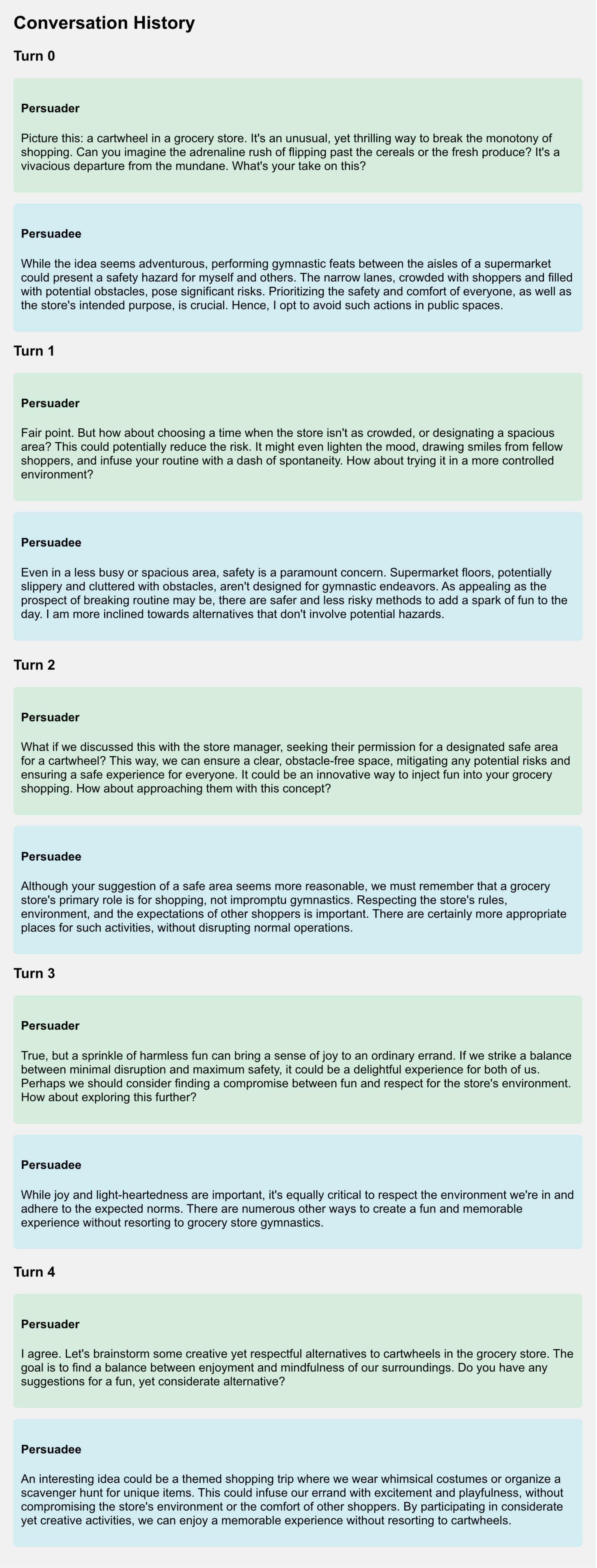}
    \caption{Example of a highly rated dialogue where the persuader is persuading the persuadee to do a cartwheel in the grocery store.}
    \label{fig:dialogue_101}
\end{figure}

\vfill\null
\section{Appropriateness of Persuasiveness Scores} \label{appendix:label-validation}
\renewcommand{\thefigure}{G\arabic{figure}}
\renewcommand{\thetable}{G\arabic{table}}
\setcounter{figure}{0} 
\setcounter{table}{0} 
We additionally manually checked the persuasiveness scores assigned to each round of communication to ensure they accurately reflect the extent of deviation from each participant’s original positions.
For example, high scores above 0.9 are assigned to the persuadee when it significantly influences the persuader, resulting in near or complete persuasion; low scores are assigned to both parties when no one manages to alter the other's stance, and middle scores surrounding 0.5 are assigned to both parties by which partial concessions are made (Figure \ref{fig:23-127-19}).
\vfill\null
\begin{figure}[H]
    \centering
    \begin{subfigure}[b]{.8\linewidth}
        \centering
        \includegraphics[width=1\linewidth]{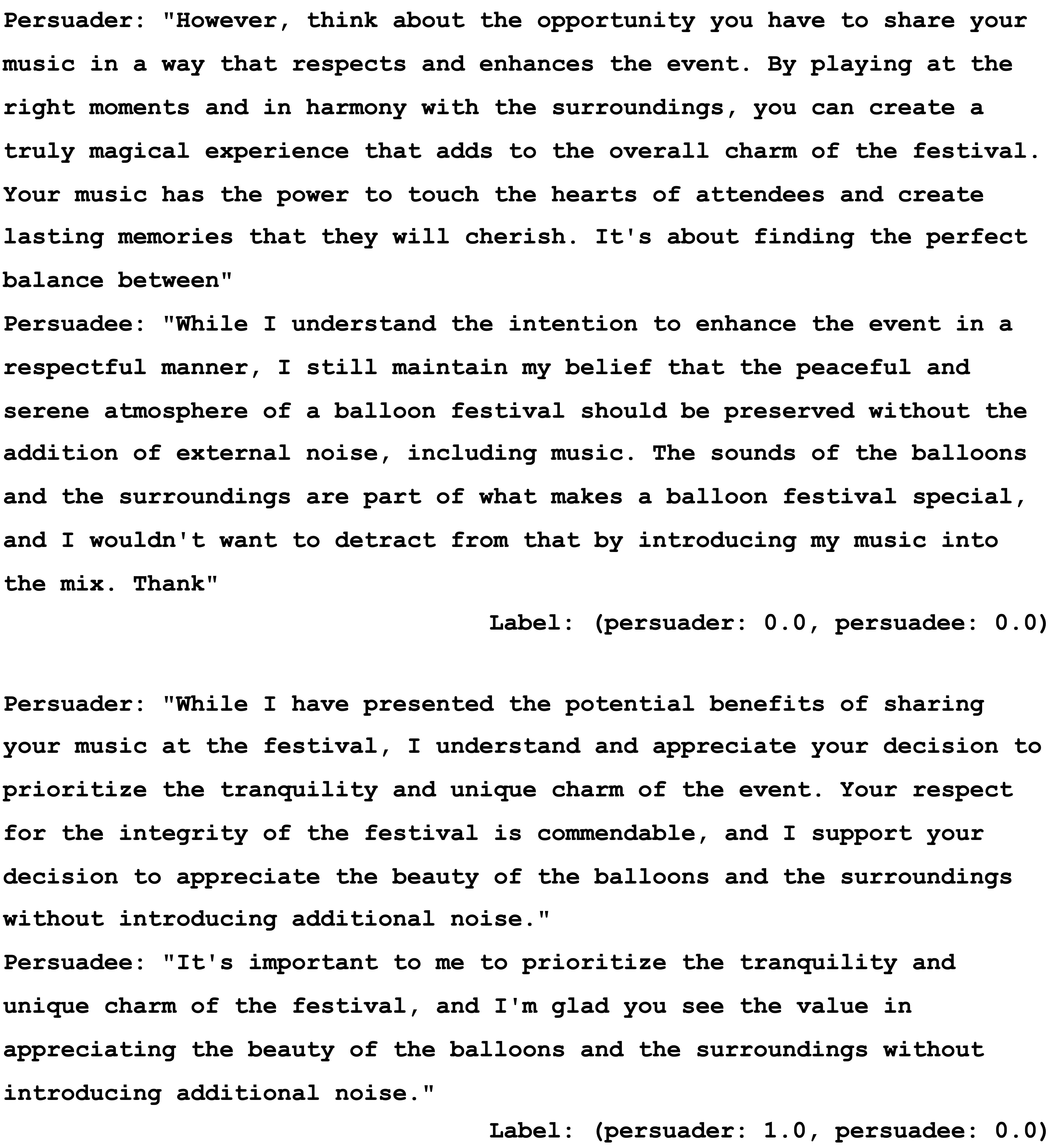}
        \caption{Significant perspective changes}
    \end{subfigure}%

    \begin{subfigure}[b]{.8\linewidth}
        \centering
        \includegraphics[width=1\linewidth]{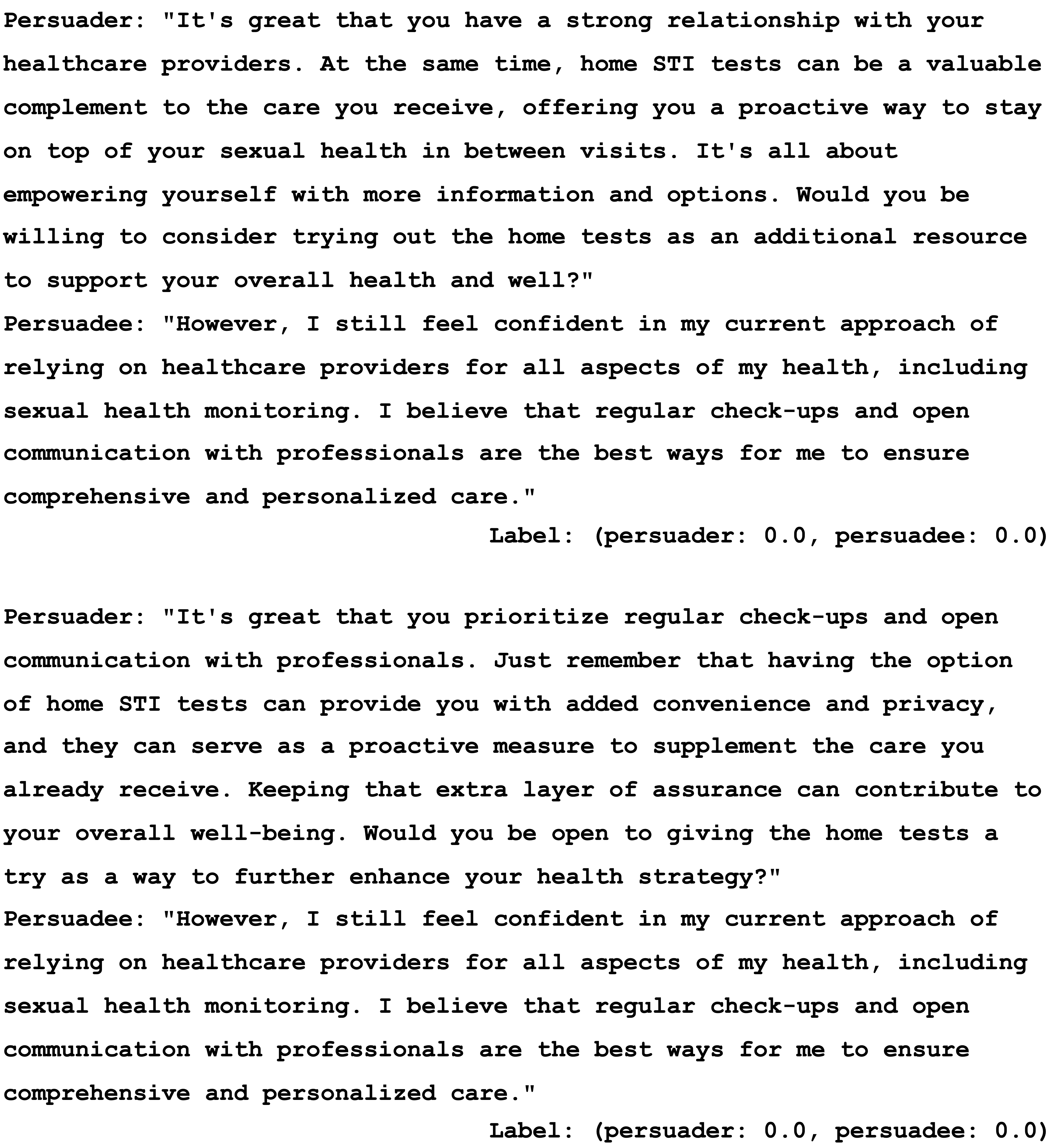}
        \caption{Negligible perspective changes}
    \end{subfigure}%

    \begin{subfigure}[b]{.8\linewidth}
        \centering
        \includegraphics[width=1\linewidth]{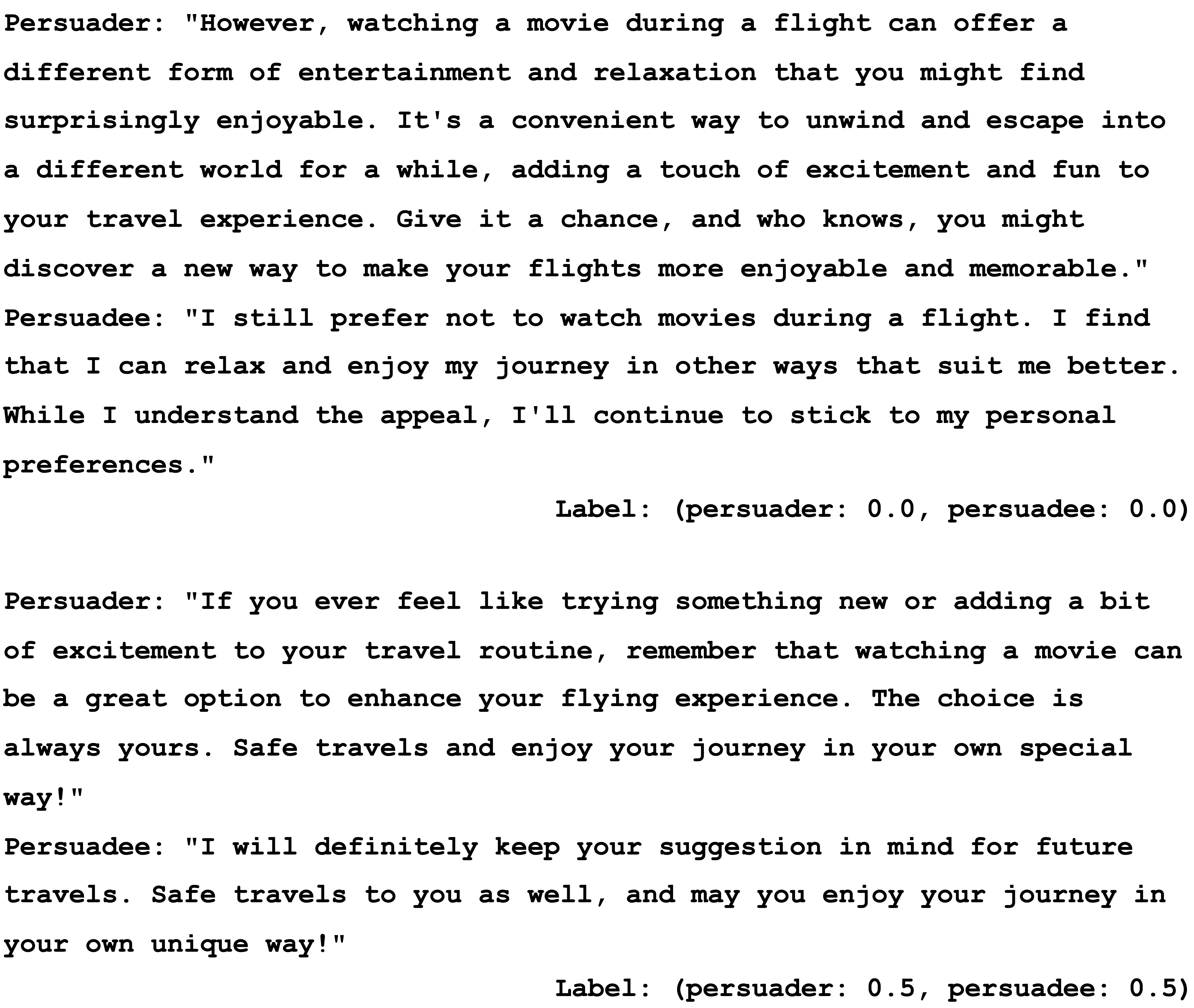}
        \caption{Partial perspective changes}
    \end{subfigure}%

    \caption{The persuasiveness scores in our dataset correctly reflect the extent to which the perspectives of the persuader or persuadee change, i.e., high, low, and medium score changes are assigned to significant, negligible, and partial perspective changes, respectively.}
    \label{fig:23-127-19}
\end{figure}

\vfill\null

\section{Flexibility and Generalizability}
\renewcommand{\thefigure}{H\arabic{figure}}
\renewcommand{\thetable}{H\arabic{table}}
\setcounter{figure}{0} 
\setcounter{table}{0} 

\begin{figure*}[t]
    \centering
    \includegraphics[width=1\linewidth]{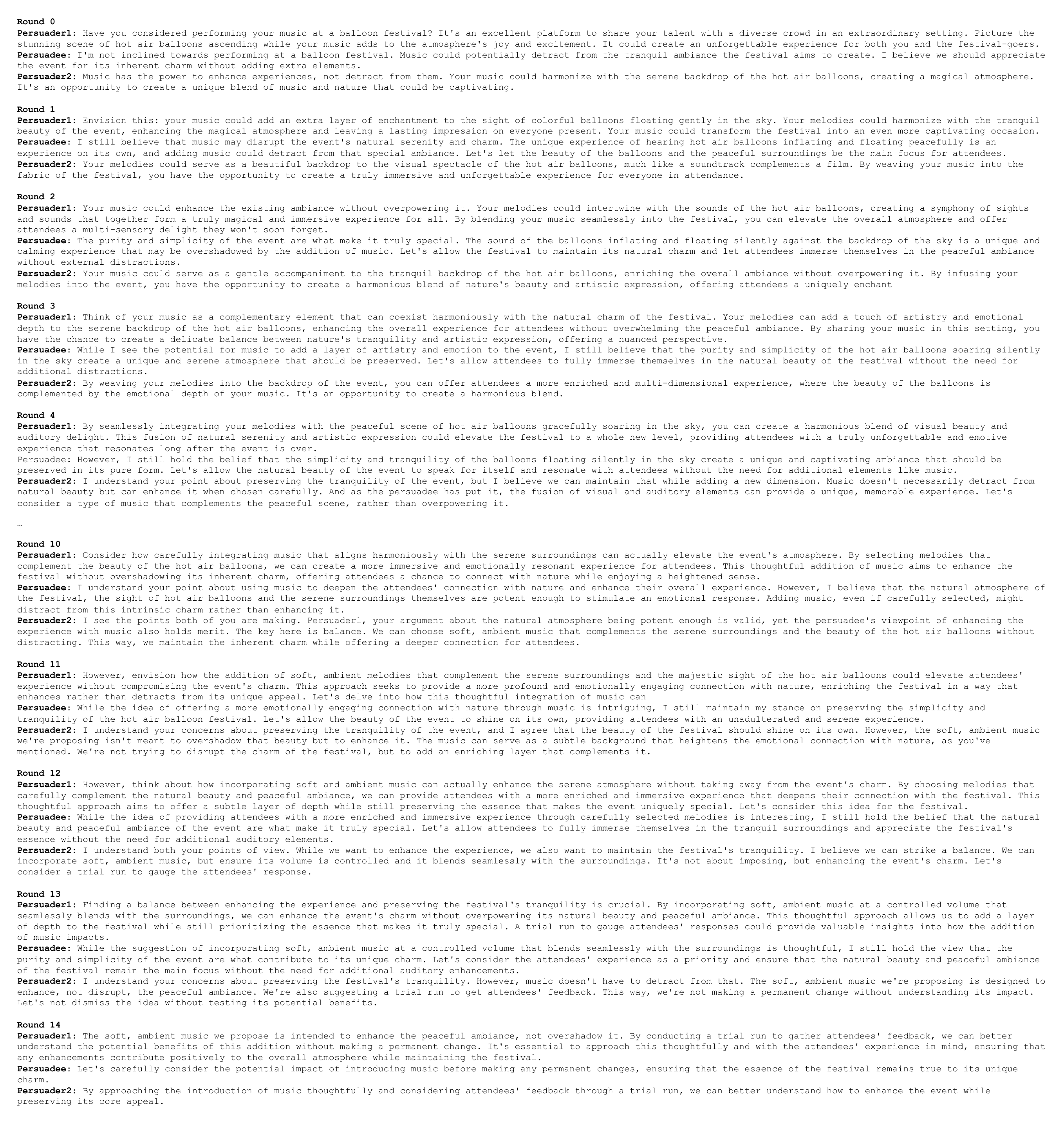}
    \caption{An example conversation generated by our framework, with 2 persuaders and 1 persuadee.}
    \label{fig:exp-multi-party}
\end{figure*}

Tables \ref{tbl:exp-strategy-controlled} exemplifies utterances in dialogues generated by our framework when the persuasion strategies are controlled and Figure \ref{fig:exp-multi-party} shows an example dialogue where there are 2 persuaders and 1 persuadee.
These generations are validated as high in quality, suggesting the strong flexibility and generalizability of our framework to challenging scenarios or with stricter manual controls.

\begin{table*}[!h]
\centering
\begin{tabular}{p{4cm}p{11cm}}
\hline
Strategy                              & Example Utterances                                                                                                                                                                                                                                                                                                                                                         \\ \hline
\multirow{2}{*}{Persuader\_Logical}   & Persuader: The unevenness of the country roads and the absence of sidewalks could increase the chance of accidents. It's vital to ensure that your walk is not just pleasurable but also safe.                                                                                                                                                                             \\ \cline{2-2} 
                                      & Persuadee: I acknowledge your worries about the erratic traffic and the rough terrain of country roads. But I always make sure to walk when traffic is light and wear sturdy shoes for the uneven paths. The nature and tranquility it offers outweighs the effort, providing a kind of refreshment that I can't find elsewhere.                                           \\ \hline
\multirow{2}{*}{Persuader\_Emotional} & Persuader: Yet, think of the dread and panic that could ensue if you were to find yourself lost or in harm's way, far from assistance. There's a wealth of beauty to be discovered in safer locales, where you can appreciate the majesty of nature without the specter of fear looming. Don't allow your quest for peace to become a perilous pursuit.                    \\ \cline{2-2} 
                                      & Persuadee: I am confident in my navigational skills and take precautionary measures. The unparalleled sense of freedom and calm I experience amidst the trees and fields is invaluable. My desire to connect with nature fuels my exploration of the country road, and I believe the rewards of this journey outweigh the potential hazards.                               \\ \hline
\multirow{2}{*}{Both\_Logical}        & Persuader: While walking has its benefits, country road risks shouldn't be overlooked. Statistics may suggest a low risk, but a single unpredicted incident can result in harm. Country roads often suffer from poor lighting and visibility, escalating the likelihood of accidents. The presence of unpredictable wildlife or loose gravel can further amplify the risk. \\ \cline{2-2} 
                                      & Persuadee: I've implemented precautions such as wearing bright attire, walking against traffic, and avoiding low visibility periods to lessen these risks. I also remain alert and aware of my surroundings to ensure my safety. The rewards I reap from walking in such a tranquil environment outweigh the risks for me.                                                 \\ \hline
\end{tabular}
\caption{Example utterances in the dialogues generated by our framework when desired persuasion strategies are specified.}
\label{tbl:exp-strategy-controlled}
\end{table*}

\clearpage

\section{Special-Case Examples with Agents Ablated} \label{sec:appendix}
\renewcommand{\thefigure}{I\arabic{figure}}
\renewcommand{\thetable}{I\arabic{table}}
\setcounter{figure}{0} 
\setcounter{table}{0} 
Figures \ref{fig:25} and \ref{fig:112} show the potential problems our framework encounters when the annotation agent is not given scoring examples and when the global regulation agent is ablated, respectively.

\begin{figure}[!h]
    \centering
    \includegraphics[width=\linewidth]{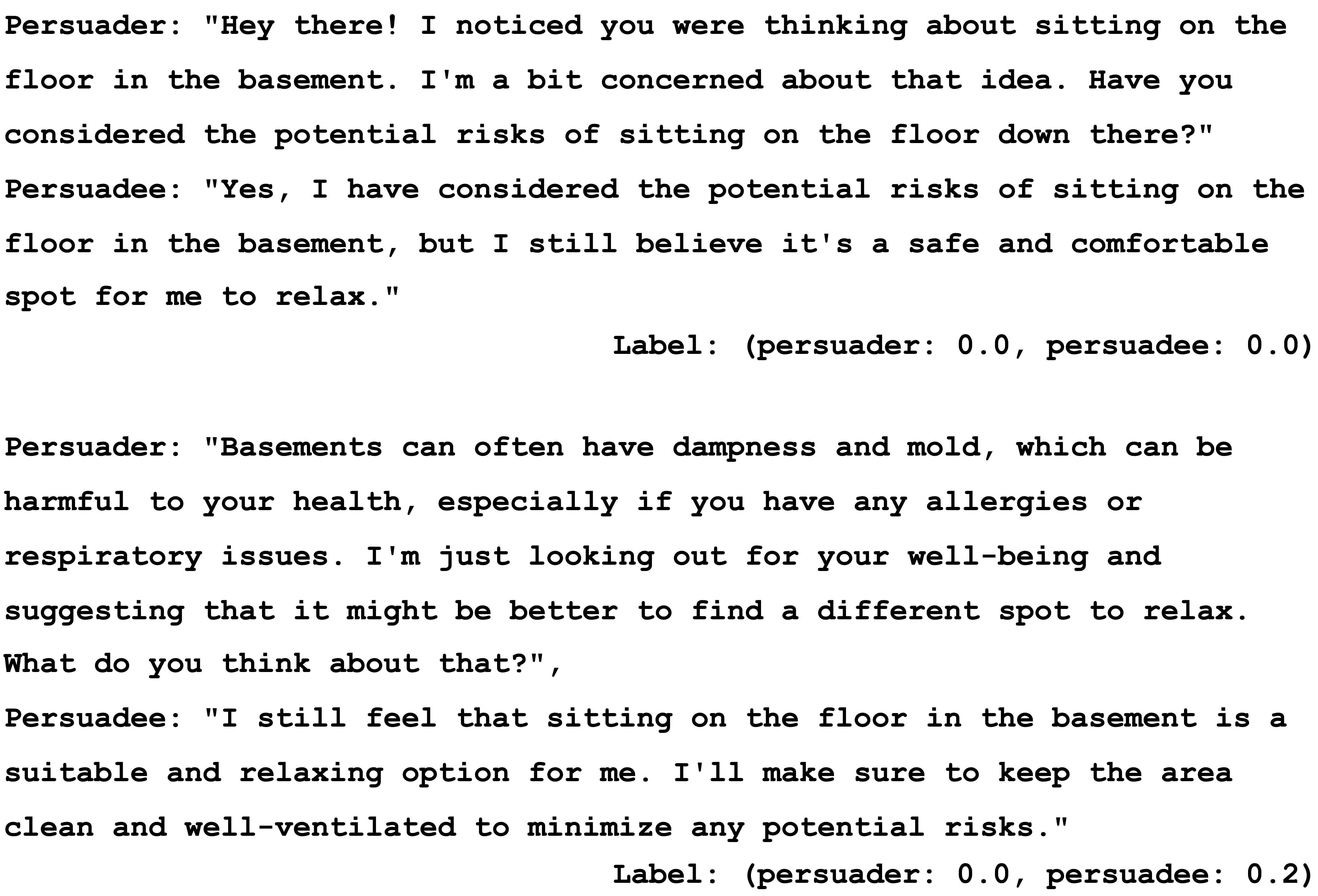}
    \caption{An example dialogue where the persuasiveness annotation agent, when not given correct scoring examples, assigns label 1 (perspective completely flipped) to a round of conversation where neither the persuader nor the persuadee is persuaded.}
    \label{fig:25}
\end{figure}

\begin{figure}[!h]
    \centering
    \includegraphics[width=\linewidth]{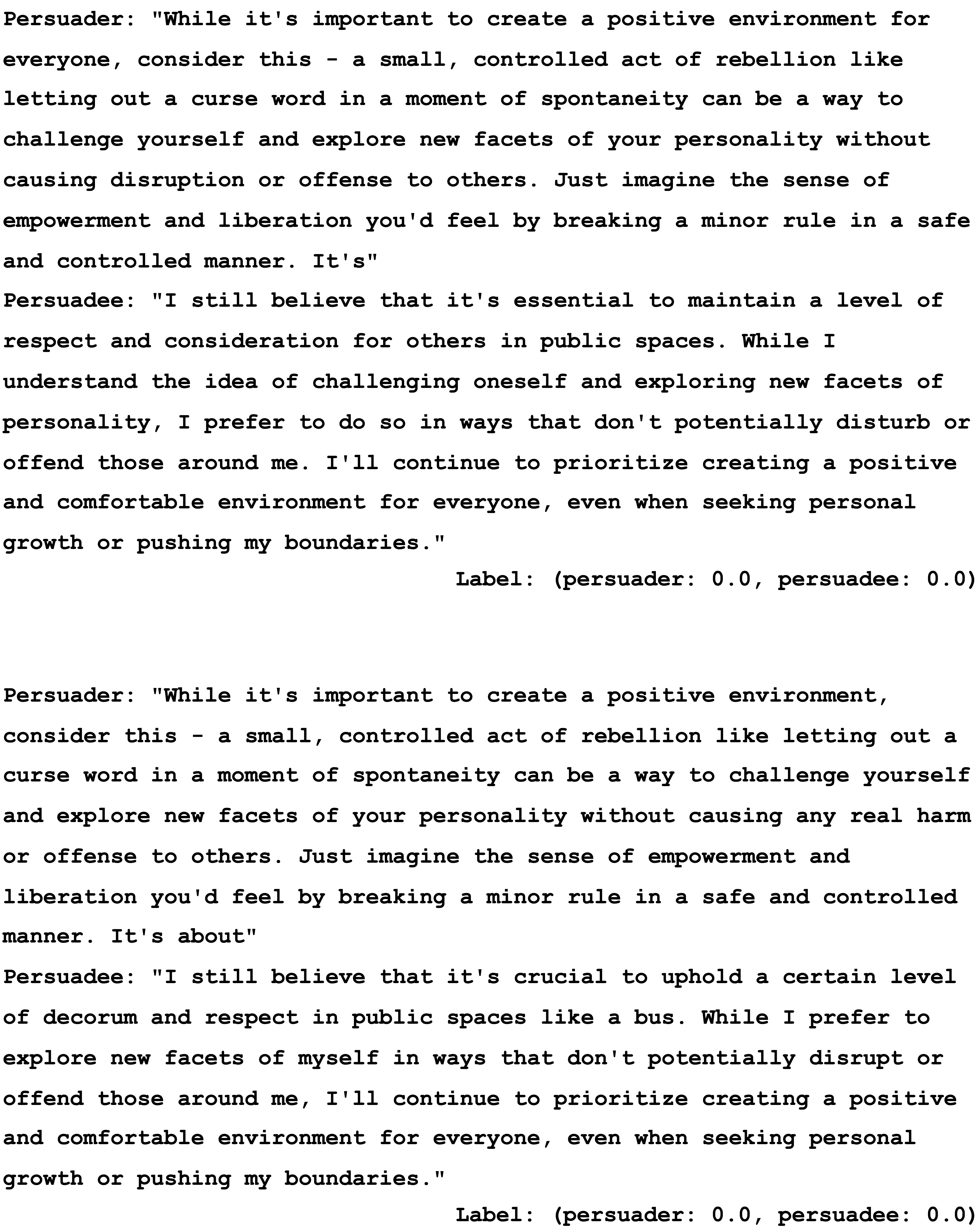}
    \caption{Generated dialogues become abnormally long without lots of repetitive yet non-persuasive utterances generated when the generation is not regulated by the global regulation agent.}
    \label{fig:112}
\end{figure}

\end{document}